\documentclass[journal]{IEEEtran}
\IEEEoverridecommandlockouts
\usepackage{cite}
\usepackage{verbatim}
\usepackage{amsmath,amssymb,amsfonts,bm}
\usepackage[linesnumbered,ruled,vlined]{algorithm2e}
\usepackage{enumitem}
\usepackage{amsthm}
\usepackage{algpseudocode}
\usepackage{textcomp}

\usepackage{float}
\usepackage[font=small]{caption}
\usepackage{xcolor}
\usepackage{wrapfig}
\usepackage{hyperref}
\usepackage[font=footnotesize,labelformat=simple]{subcaption}
\usepackage[protrusion=true,expansion=true]{microtype}
\pdfoutput=1

\usepackage{caption}
\usepackage{multirow}
\usepackage{graphicx}
\usepackage{cleveref}

\def\BibTeX{{\rm B\kern-.05em{\sc i\kern-.025em b}\kern-.08em
    T\kern-.1667em\lower.7ex\hbox{E}\kern-.125emX}}
    
    \makeatletter 
\newcommand\semiHuge{\@setfontsize\semiHuge{22.7}{28.38}}
\makeatother

\begin{document}

% \setcounter{secnumdepth}{4}

% Cooperative Federated Unsupervised Learning: Towards Smart Data Exchange For Distributed Learning

\title{Unsupervised Federated Optimization at the Edge: D2D-Enabled Learning without Labels
}

\author{
\IEEEauthorblockN{Satyavrat Wagle\IEEEauthorrefmark{1}, Seyyedali Hosseinalipour\IEEEauthorrefmark{2}, \\Naji Khosravan\IEEEauthorrefmark{3}, and Christopher G. Brinton\IEEEauthorrefmark{1}
\thanks{This paper is an extension of the conference paper \cite{ufl} which appeared in the 2022 IEEE Global Communications Conference.}
}
\IEEEauthorblockA{\IEEEauthorrefmark{1}School of Electrical and Computer Engineering, Purdue University, West Lafayette, IN, USA}
\IEEEauthorblockA{\IEEEauthorrefmark{2}University at Buffalo-SUNY, Buffalo, NY, USA}
\IEEEauthorblockA{\IEEEauthorrefmark{3}Zillow Group, Seattle, WA, USA}
\IEEEauthorblockA{\IEEEauthorrefmark{1}\{wagles, cgb\}@purdue.edu, 
\IEEEauthorrefmark{2}alipour@buffalo.edu,
\IEEEauthorrefmark{3}najik@zillowgroup.com}%\vspace{-8mm}
}
%}
% \author{\IEEEauthorblockN{1\textsuperscript{st} Given Name Surname}
% \IEEEauthorblockA{\textit{dept. name of organization (of Aff.)} \\
% \textit{name of organization (of Aff.)}\\
% City, Country \\
% email address or ORCID}
% \and
% \IEEEauthorblockN{2\textsuperscript{nd} Given Name Surname}
% \IEEEauthorblockA{\textit{dept. name of organization (of Aff.)} \\
% \textit{name of organization (of Aff.)}\\
% City, Country \\
% email address or ORCID}
% \and 
% \IEEEauthorblockN{2\textsuperscript{nd} Given Name Surname}
% \IEEEauthorblockA{\textit{dept. name of organization (of Aff.)} \\
% \textit{name of organization (of Aff.)}\\
% City, Country \\
% email address or ORCID}
% \and
% \IEEEauthorblockN{2\textsuperscript{nd} Given Name Surname}
% \IEEEauthorblockA{\textit{dept. name of organization (of Aff.)} \\
% \textit{name of organization (of Aff.)}\\
% City, Country \\
% email address or ORCID}
% \and
% \IEEEauthorblockN{2\textsuperscript{nd} Given Name Surname}
% \IEEEauthorblockA{\textit{dept. name of organization (of Aff.)} \\
% \textit{name o9f organization (of Aff.)}\\
% City, Country \\
% email address or ORCID}
% }

\maketitle

\thispagestyle{plain}
\pagestyle{plain}
% textwidth = 8.85 cm

\begin{abstract}
Federated learning (FL) is a popular solution for distributed machine learning (ML). While FL has traditionally been studied for supervised ML tasks, in many applications, it is impractical to assume availability of labeled data across devices. To this end, we develop Cooperative Federated unsupervised Contrastive Learning ({\tt CF-CL)} to facilitate FL across edge devices with unlabeled datasets. {\tt CF-CL} employs local device cooperation where either explicit (i.e., raw data) or implicit (i.e., embeddings) information is exchanged through device-to-device (D2D) communications to improve local diversity. 
Specifically, we introduce a \textit{smart information push-pull} methodology for data/embedding exchange tailored to FL settings with either soft or strict data privacy restrictions. Information sharing is conducted through a probabilistic importance sampling technique at receivers leveraging a carefully crafted reserve dataset provided by transmitters. In the implicit case, embedding exchange is further integrated into the local ML training at the devices via a regularization term incorporated into the contrastive loss, augmented with a dynamic contrastive margin to adjust the volume of latent space explored. Numerical evaluations demonstrate that {\tt CF-CL} leads to alignment of latent spaces learned across devices, results in faster and more efficient global model training, and is effective in extreme non-i.i.d. data distribution settings across devices.

% {\tt CF-CL} introduces a push-pull smart data sharing  mechanism tailored for unsupervised settings, where a subset of local dataset is initially pushed to the neighboring devices and used as reserved data points (anchors), based on which data pulls are carried out based on importance estimation.
% To facilitate this, we introduce a set of new data sampling and exchange techniques as well as embedding integration methods for {\tt CF-CL}. 
% Numerical evaluations on real-world ML task datasets demonstrate the superiority of {\tt CF-CL} as compared to state-of-the-art methods. 
%
% \cgb{Notion of ``contrastive'' should be brought out here?}
\end{abstract}

% \begin{IEEEkeywords}
% federated learning, unsupervised learning, contrastive learning, importance sampling
% \end{IEEEkeywords}
% \vspace{-5mm}

\vspace{-1mm}
\section{Introduction}\label{intro}
% \noindent=      q34567890-=-0987657890-09876890-98789098768909878
Many emerging intelligence tasks require training machine learning (ML) models on a distributed dataset across a collection of wireless edge devices (e.g., smartphones, smart cars) \cite{fi11040094}. Federated learning (FL) \cite{9084352}
%,MAL-083} 
utilizes the computational resources available at edge devices for data processing. Under conventional FL, model training consists of (i) a sequence of local iterations by devices on their individual datasets, and (ii) periodic global aggregations by a main server to generate a global model that is synchronized across devices to begin the next training round.
%which is then broadcast across them to initiate the next ML model training round.

In this work, we are motivated by two fundamental challenges related to implementation of FL over real-world edge networks. First, device datasets are often non-independent and identically distributed (non-i.i.d.), causing local models to be biased and a significant degradation in global model performance. Second,  data samples collected by each device (e.g., images, sensor measurements) are often unlabeled, which makes supervised ML model training impossible. We aim to jointly address these challenges with a novel methodology for information sampling and exchange across devices while remaining sensitive to any data privacy restrictions.
%blend latent embedding techniques from ML with smart information sharing among devices for  FL.

%FL trains an ML model via a sequence of 
%local ML model training rounds at the devices using their local datasets. The local models of the devices are periodically aggregated at a server to generate a global model, which is broadcast across them to initiate the next ML model training round. Implementation of FL over real-world networks faces a number of challenges, two of which are of our particular interest: (i) datasets of the devices are non-independent and identically distributed (non-i.i.d.), and thus their local models can get biased toward their local datasets, which can significantly degrade the global model performance; (ii) data available/collected at the devices is often unlabelled. The former challenge has been studied extensively in literature 
%\cite{DBLP:journals/corr/abs-1806-00582}, \cite{8889996},\cite{DBLP:journals/corr/abs-2006-09637}. However, the latter challenge is widely underexplored in the current literature~\cite{furl},~\cite{Li_2021_CVPR}.
%has been mostly neglected and there is only a few recent works devoted to investigating that \ali{[CITE]}.

\vspace{-1mm}
\subsection{FL with Labeled vs. Unlabeled Data}
% \ali{Here we need to first clear that what applications have labelled data and what applications do not! Then we say many applications do not have unlabelled data. It is costly to obtain labels.}
{

%  In classical FL \cite{fedavg}, it is assumed that a well curated/supervised set of labeled data is available at the end devices. 
Most works in FL assume labeled data across the devices~\cite{fedavg}.
%,hard2018federated}.
Large amounts of unlabeled data can be useful to learn representations of a large scale distributed dataset with the objective of adapting to many downstream tasks, such as classification, segmentation, etc., through zero or few shot learning~\cite{9589192}.
%Most of the current 
%However, in many real-world environments, the IoT edge devices (e.g., distributed sensor networks, smart cars, smartphones), generate very large amounts of data \cite{iot_bigdata}, labelling of which may be impractical \cite{active}.
%For example, the images taken by the cameras mounted on smart cars, are mostly unlabeled and it is not clear whether the picture taken is a stop sign, a speed limit sign, or a pedestrian crosswalk sign.
%Furthermore, many
%applications such as speech separation \cite{9632783}, image segmentation for facial recognition \cite{feddis} and domain adaptation \cite{fedfr}, which highly desire to utilize FL due to the distributed nature of data collection at the network edge, are often faced with unlabeled data at the end users which would pose significant challenges with regard to label curation \cite{active} and privacy concerns \cite{Mothukuri_Parizi_Pouriyeh_Huang_Dehghantanha_Srivastava_2021}. 
%Although training on unlabeled data (i.e., unsupervised training) is a more practical setting for FL, majority of the current literature are focused on FL implementation over labeled datasets at the end devices (i.e., supervised training). This is mostly due to the tractable theoretical analysis that supervised learning offers, which can be exploited to conduct system-level design. 
In the following, 
we provide a brief literature review to summarize the extensive line of work in supervised FL, and then detail the current state of the art in unsupervised FL.}
%We next provide an overview on the literature on FL focusing on the two aforementioned challenges.
%While the former challenge has received considerable attention in recent literature \cite{DBLP:journals/corr/abs-1806-00582,8889996,DBLP:journals/corr/abs-2006-09637}, the latter is widely underexplored~\cite{furl,Li_2021_CVPR}.

\subsubsection{FL under labeled data}
Researchers have aimed to address the impact of non-i.i.d. device data distributions on FL performance. In~\cite{wang2019adaptive},
%9562522}
convergence analysis of FL via device gradient diversity-based metrics is conducted, and control algorithms for adapting system parameters (e.g., aggregation frequencies) to optimize convergence speeds are proposed.
%In \cite{DBLP:journals/corr/abs-2010-15582}, probable causes of performance degradation in non-i.i.d. settings is discussed and countermeasure methods such as device side gradient projection are proposed.
In \cite{wang2020optimizing}, a reinforcement learning-based method for device selection is introduced, aiming to counteract local model biases caused by non-i.i.d. data. In~\cite{briggs2020federated}, a clustering-based approach is developed, aiming to produce a hierarchy of local models that captures their diversity. In~\cite{zhao2021federated}, the authors tune the global model aggregation procedure to reduce local model divergence using a theoretical upper bound.
%The authors of \cite{DBLP:journals/corr/abs-2102-02079} characterize data partitioning strategies to reduce the effect of non-i.i.d. data on the performance of FL.

% Works such as \cite{9311906,wang2021device,furl,zhao2018federated} explored data exchange between devices in FL to improve local data similarities in settings with no strict privacy concerns on data sharing.

%This has motivated recent works on FL for unlabeled data.
%this is not Although the above works provide valuable insights, they consider labeled data at the devices (i.e., supervised ML), which is not practical in many scenarios. 

\subsubsection{FL under unlabeled data}
A few works have recently considered unsupervised FL. In \cite{berlo}, a local pretraining methodology was introduced to generate unsupervised device model representations for downstream machine learning tasks. The authors in \cite{furl} proposed addressing the inconsistency of local representations arising from non-i.i.d unlabelled datasets through a  dictionary-based method implemented at the server. In \cite{9348203}, the dataset imbalance problem is addressed with weight aggregation at the server being defined according to inferred sample densities. In \cite{unsupfl}, unsupervised FL is considered for the case where unlabeled device data has been subdivided into sets that can be treated as surrogate labels for training. The authors in  \cite{Li_2021_CVPR} exploit similarities across locally trained unsupervised model representations to correct against biases by maximizing local agreement with the global models. 
%through a server aggregation mechanism with weights fixed size bins along with self organizing maps are used  to generate soft labels and devices are weighed based on 
%There exist a few works on FL with unlabeled data across the devices (i.e., unsupervised ML), where battling the impact of non-i.i.d. data on ML model training remains an important challenge. 
%In \cite{DBLP:journals/corr/abs-2109-07504}, exchange of representational metadata in the form of dictionaries is used to enhance the model training. 

In this work, we consider \textit{contrastive learning} (CL) as our framework for unsupervised ML, and extend it to the federated setting.
%\cite{1640964,
CL is an ML technique which aims to learn embeddings of unlabeled datapoints that maximize the distance between different data points and minimizes it for similar points in the latent space \cite{pmlr-v119-chen20j}. Our work adapts contrastive learning to a federated setting, while avoiding the additional work of assigning surrogate labels to datapoints such as in \cite{unsupfl}. Our work enables contrastive learning to be performed on exclusively unlabelled data, as opposed to a combination of supervised and contrastive learning, such as in \cite{Li_2021_CVPR}. 
%\cgb{Edited} 
Along these lines, in \cite{zhuang2022divergenceaware}, the authors propose a self-supervised learning framework for federated systems using contrastive loss, and propose a divergence-aware method for local model updates which adjusts local models towards global knowledge. In contrast, our work employs the contrastive learning framework to identify and share crucial information between devices, to promotes alignment between local models during federated training.

% In FL, contrastive methods such as \cite{furl,Li_2021_CVPR} utilize similarity metrics for alignment of remote models across devices.

\subsection{FL with Information Exchange Among Devices}

One of the major challenges faced in the implementation of FL over real-world networks is data heterogeneity across devices, which as discussed in Sec. I-A, leads to local model bias. A line of solutions to this challenge for settings that are not completely privacy restricted is to enable some form of \textit{information} exchange among the devices.

%\cgb{Edited} 
When information sharing is permitted in FL, there are many scenarios in which inter-device information exchange may be preferred to sharing the information with a server and training the model in a centralized manner. This includes settings where the server is far away from the devices (e.g., in large wireless cells) \cite{r32_sdist1}, where there is a large hierarchy of nodes separating device and server (e.g., in hierarchical fog networks) \cite{9705093}, and/or where the server's computational capabilities beyond facilitating model aggregations are limited \cite{server_relay1,server_relay2}. In such settings, centralization may be prone to large energy consumption and/or time delays compared with periodically sharing small amounts of information over local device networks \cite{wang2023towards}.

%For example, in agricultural IoT systems where devices are typically located in farmlands, while servers arelocated much further away \cite{r32_sdist1}, or energy constrained systems such as wireless sensor networks \cite{auto_offloading} where small battery powered devices are expected to operate for a long duration of time.}

% \ali{Here, we talk about the fact that we have two types of information: Data and XXXX (come up with a good name, embedding is not good since it only applies to our problem. Maybe ``explicit/raw information and ``implicit/processed information""), and then talk about one of them can be applied in applications where they do not require data privacy (give 3 examples). then talk about the other one that can be applied in applications where they need data privacy (give 3 applications).
% Here we can mentioned the prior works that have tried some notation of information exchange among the devices. mention that this is an emerging area, which there are many open problems. Also, throughout the paper, you can keep adding ``Remarks" and talk about your ideas in each part for future work.}
Information can be classified into two categories: (i) \textit{explicit} or \textit{raw information}, which contains the full set of features describing the datapoints (e.g., RGB colors of the pixels in a photo), and (ii) \textit{implicit} or \textit{encoded} information, which correspond to processed datapoint features represented in a lower dimensional space upon applying some form of filtration (e.g., embeddings or projections). 

In a networked system, explicit information exchange exposes raw user information, which makes it suitable only for applications that are not subject to stringent restrictions of data privacy, such as computation offloading in (i) wireless sensor networks \cite{wsn_survey}
%,\cite{drl_multiuser_wsn}, 
%\cite{auto_offloading}, 
(ii) self driving vehicles \cite{eff_veh},
%\cite{delay_veh},
%\cite{sd_iov} 
and (iii) resource constrained IoT devices \cite{learning_iot}
%\cite{hier_iot}
. One of the main advantages of explicit information exchange in a FL setting is that is does not require any pre-processing of features of data, and thus makes it suitable for resource constrained edge devices to incorporate into training (e.g., sensors, and unmanned aerial vehicles). On the other hand, in applications with strict data privacy concerns (e.g. facial recognition \cite{feddis},
%\cite{fedfr},
user activity recognition \cite{user_rec}, and some text mining applications \cite{nlp_fl}), exchange of only implicit information is permitted, since it is retrieving raw data features from the encoded information is extremely difficult ~\cite{azam2022can}. That being said, obtaining implicit/encoded information requires pre-processing of data at the end users, which in turn can impose high latency and computation burden. 

Information exchange among FL devices has been studied by a few recent works, most of which are concerned with supervised FL~\cite{wang_devicesampling,zhao2018federated,furl,9311906}. In this work, we aim to design smart data (i.e., explicit information) and embedding (i.e., implicit information) exchange for cooperative FL systems via exploiting device-to-device (D2D) communications. %\cgb{Edited}
Our method can be applied for scenarios which allow for inter-device communication, and is compatible with different levels of privacy restrictions. We employ explicit information exchange for scenarios which are not subject to stringent privacy restrictions, such as wireless sensor networks aiming to learn about an environment across different modalities of data \cite{explicit_allowed1}, vehicle to vehicle (V2V) networks which benefit from exchanging traffic information among vehicles to optimally plan routes \cite{r32_ds2}, and swarm robotics where agents aim to cooperate to achieve a common goal \cite{explicit_allowed2}. On the other hand, for scenarios where data privacy is a primary concern, such as distributed healthcare systems
\cite{r33_sens1},
%,r33_sens2,r33_sens3}
we employ implicit information exchange, with embeddings that cannot be used to reconstruct source datapoints. We also aim for our method to facilitate information exchange in a fully distributed manner, i.e., without any added control overhead at the FL server.

In the following, we provide a review of works concerned with information exchange in FL, and highlight the current art in unsupervised FL.

% applicable where locally available data is sensitive and requires a higher degree of privacy, such as facial recognition \cite{feddis},\cite{fedfr}, user activity recognition \cite{user_rec} or text prediction for smartphones \cite{nlp_fl}.

% \ali{}

\subsubsection{FL with explicit information exchange} Recent in supervised FL have shown that, when permissible, even a small amount of raw data exchange among devices can substantially improve training~\cite{wang_devicesampling,zhao2018federated,furl,9311906}.
However, to the best of our knowledge, the impact of raw data sharing in unsupervised FL has not been studied. In contrastive learning, the model at each device can use exchanged information as negatives for anchors selected from local data, and as anchors where negatives are selected from local data to improve the overall training performance (see Sec.~\ref{sec:ML} for the definition of negative and anchor datapoints). To exploit D2D communications as a substrate for improving local model alignment, we design a probabilistic smart data push-pull strategy for unsupervised FL that enables devices to exchange raw datapoints that have the highest expected impact on the performance of the global model.

% The process of FL using explicit information is defined in Algorithm \ref{alg:flde} in step \ref{state:update_local_model_exp}.

\subsubsection{FL with implicit information exchange}\label{sub:implicit_exchange_fl}

Implicit information exchange has been recently studied by a few works for unsupervised and semi-supervised FL. The authors in \cite{furl} have attempted to address the alignment of representations in unsupervised FL by incorporating implicit information exchange. However, this method requires the availability of a \textit{public dataset} at the server, datapoints of which are sampled from the devices. Availability of public data is a common practice in personalized FL, where a public dataset is used for knowledge distillation \cite{dd_fl},\cite{feddf} and transfer learning \cite{fedmd}. 
%\cgb{Edited} 
However, reliance on such a dataset renders the method of \cite{furl} not directly applicable to our implicit information exchange setting where raw data sharing is not allowed. Alternatively, methods such as \cite{fedcon} attempt to align the local models of the devices with a classification model learned at the server. Such methods rely on availability of labelled data, as well as a trainable model at the server.

%Implicit information exchange has been recently studied by a few works for unsupervised and semi-supervised FL. The authors in \cite{furl} have attempted to address the alignment of representations in unsupervised FL by incorporating implicit information exchange. However, this method requires the availability of a \textit{public dataset} at the server, datapoints of which are sampled from the devices. Availability of public data is a common practice in personalized FL, where a public dataset is used for knowledge distillation \cite{dd_fl},\cite{feddf} and transfer learning \cite{fedmd}. However, reliance on such a dataset renders the method of \cite{furl} not applicable to setting with strict privacy concerns where raw data sharing is not allowed.  Alternatively, methods such as \cite{fedcon} attempt to align the local models of the devices with a classification model learned at the server. However, this method also relies of availability of labelled data as well as a trainable model at the server.

% \ali{Methods such as \cite{furl} use methods that centralize the process of information exchange by initiating the information exchange at time-steps where model aggregation is conducted. except centralization let us say sth else as well... what is their implicit info?}

As compared to the above works, we study a different problem, which aims to conduct smart implicit information exchange among the devices via D2D communications for unsupervised FL. 
%\cgb{Edited} 
We aim to facilitate the information exchange process in a distributed manner, without introducing any control overhead for a main server or alternative aggregation point.

%As compared to the above works, we study a different problem, which aims to conduct smart implicit information exchange among the devices via D2D communications for unsupervised FL. We use data embeddings generated via the local models as implicit information and incorporate it into the local training at the devices. Our technique introduces a regularization term into the conventional definition of contrastive loss used in literature, which enables exploiting the received embeddings to \textit{debias} the local model while taking into account for the \textit{freshness} of the received embeddings from the neighboring devices. 

% In contrastive learning, we modify the loss function defined in \ref{eqn : triplet_loss}, by adding a proximal loss based on the received implicit information. The process of FL using implicit information is defined in Algorithm \ref{alg:flde} in step \ref{state:update_local_model_imp}

\subsection{Implicit and Explicit Information Exchange}\label{sub:diff}
We characterize explicit information exchange as an exchange of raw datapoints, and implicit information exchange as an exchange of embeddings of datapoints generated by local models. Properties unique to each of these exchange modalities lend themselves to certain system characteristics. Explicit information is ground truth, which is immutable, and hence does not suffer from ``staleness.'' Implicit information on the other hand, is transient, as it is a function of local model $\phi_i^t$ of device $i$ at time-step $t$. The fidelity of information received at time $t$ reduces with time, thus necessitating more frequent information exchange. However, implicit information encodes the data by the local model $\phi_i^t$, making it less of an exposure as compared to explicit information. Implicit information exchange also provides users with flexibility in choosing their communication payload, as the dimensionality of the representation is a user defined parameter. The size of explicit information, on the other hand, is predetermined. In case of implicit information exchange, adjusting this dimension allows us to improve communication and computation requirements for resource-limited devices.

Thus, explicit information exchange is favored for systems working with low risk data, and where frequent exchange of information is not feasible, e.g., in the wireless sensor networks example mentioned above. On the other hand, implicit information exchange is favored for systems that deal with data in situations where privacy is one of the primary concerns, but where frequent communication with other devices in the network is possible from a time and resource consumption perspective, e.g., in the distributed healthcare system example.
%An example of this would be a network of hospitals, which make inferences on confidential patient data, but have significant compute power to facilitate frequent exchanges of obfuscated information.
%. An example of this is wireless sensor networks, which work with environmental data but are resource constrained and often geographically distant

% Further, to minimize the amount of data exchange, we develop a novel  paradigm.
% \ali{Last paragraph should be a conclusion and differentiation of your work vs. the works you mention above! }

% {Methods such as \cite{furl} use methods that centralize the process of information exchange by initiating the information exchange at time-steps where model aggregation is conducted. Our method, however allows for D2D communications, which decentralize the information exchange process, and thus decoupling the aggregation and information exchange processes. This also allows devices to take advantage of geographical proximity of devices and reduce potentially costly communication with a remote server.}

\subsection{Importance Sampling vs. Information Exchange}
The connection between our work and the notion of importance sampling in ML \cite{is_mb},\cite{is_stochastic} is worth mentioning.
In the ML community, importance sampling techniques have been introduced to accelerate training through the choice of minibatch data samples \cite{Dong_2018_ECCV}. In FL, by contrast, importance sampling has typically been employed to identify \textit{devices} whose models provide the largest improvement to the global model, e.g., \cite{9413655,wang_devicesampling}. Our work extends the literature of importance sampling to consider inter-device datapoint/embedding transfers in a federated setting, where devices exchange their local datapoints and embeddings to accelerate model training speed.

\begin{figure}[t]
    \centering
    \includegraphics[width=.95\columnwidth]{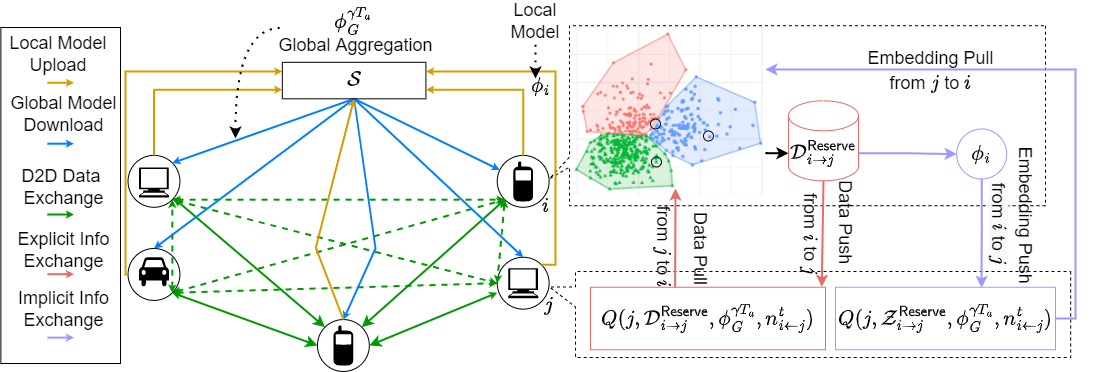}
    \vspace{-1mm}
    \caption{{\tt{CF-CL}} introduces smart push-pull information transfer to improve unsupervised FL based on importance information sampling. 
    % We use local model $\phi_i$ to calculate approximate importance of datapoints $\{d_j\}$ and selecting $\tilde{\mathcal{D}_j^t}, t = \tau T_p$ before  training $\phi_i$ on $\mathcal{D}_i \cup \tilde{\mathcal{D}_j^t}$.
    }
    \label{ss_num_clusters}
    \vspace{-0.25in}
\end{figure}

\vspace{-1mm}
\subsection{Summary of Contributions}

% \ali{use itemize and list all of them. Add the regularization stuff}
% \ali{TBW}
%\begin{itemize}
{
\begin{itemize}
    \item We develop {\tt CF-CL} -- Cooperative Federated unsupervised Contrastive Learning -- a novel method for unsupervised FL. {\tt CF-CL} improves training speed via smart D2D information exchange among the devices in a cooperative framework. {\tt CF-CL} is a general plug-and-play technique that can be mounted on current unsupervised FL methods.
    \item We introduce a novel D2D-based information push-pull strategy based on probabilistic importance sampling in {\tt CF-CL}. For explicit information exchange, we characterize the importance of a remote datapoint via a joint clustering and loss measurement technique, aiming to accelerate the convergence rate of FL in the presence of limited communication resources among the devices. We also develop a method for intelligent reserve data selection which identifies a set of datapoints that promote a balanced representation of the overall distribution to improve importance calculations.
    \item For implicit information exchange, we develop a probabilistic embedding sampling technique for {\tt CF-CL}.
    To incorporate the exchanged embeddings among the devices in the training process,  we introduce a regularization term into the conventional definition of contrastive loss which promotes model debiasing.  We also introduce a dynamic contrastive margin for exchanged implicit information to adjust the volume of latent space explored for identifying hard negatives based on cluster size.
    \item Our numerical experiments demonstrate that {\tt CF-FL} significantly improves FL training in terms of (i) alignment of unsupervised learned latent spaces, (ii) global ML model convergence, and (iii) communication resources and delay incurred to reach target accuracy levels. We find this for non-i.i.d. data distributions across devices under both explicit and implicit information sharing mechanisms.
\end{itemize}
An abridged conference version of this work appeared in \cite{ufl}. Compared to \cite{ufl}, we make the following key additions: (i) We extend {\tt{CF-CL}} to be applicable to embeddings (implicit information) in addition to datapoints (explicit information) as presented in the conference version. (ii) We adapt the \textit{smart push-pull} of information to identify important embeddings using a distance based probablistic selection mechanism. (iii) We address the inherent drawbacks of implicit information, namely staleness and consequently time-varying importance values by introducing a time-variant scaling operation which aims to reduce the influence of obsolete implicit information. (iv) We significantly expand upon our numerical section (Sec. \ref{experiments}) by conducting experiments on an additional dataset as well as analyzing the alignment of the embeddings produced by {\tt{CF-CL}}, the resource costs of {\tt{CF-CL}}, the impact of device selection per aggregation and the effect of choosing local models for importance calculations.

\vspace{-.5mm}
\section{System Model and Machine Learning Task}
% \ali{let us stick to ``device" instead of ``device"}
\noindent An overview of our method is illustrated in Fig~\ref{ss_num_clusters}. In this section, we go over our network model (Sec.~\ref{sec:net}) followed by the ML task for unsupervised FL (Sec.~\ref{sec:ML}).

%first describe our network model (Sec.~\ref{sec:net}) and then present the ML task  of unsupervised FL (Sec.~\ref{sec:ML}).
% \ali{let us stick to my notaions below}
\vspace{-1.2mm}
\subsection{Network Model of FL with Information Exchange}\label{sec:net}
% \ali{We can have the description for a general class, and call it ``Information" and use ``$\mathcal{I}$" instead of ``$\mathcal{D}$" }

 We consider a network consisting of a server $S$ and a set of devices given by the set $\mathcal{C}$. At each time-step $t$, each device $ i\in \mathcal{C}$ possesses a local ML model parametrized by $\bm{\phi}_i^t\in\mathbb{R}^p$, where $p$ is the number of model parameters. Let $\mathcal{D}_i$ denote the \textit{initial} local dataset at device $ i$, i.e., before any information exchange. Each local model $\phi_i^t$ is trained on $\mathcal{D}_i$ as well as information received from neighboring devices. The server aims to maintain a global model $\bm{\phi}_G^{t}$ via aggregating $\{\bm{\phi}_i^t\}_{i\in\mathcal{C}}$.

  We represent the D2D communication graph between the devices via $\mathcal{G}=(\mathcal{C},\mathcal{E})$ with vertex set $\mathcal{C}$ and edge set $\mathcal{E}$. The existence of an edge between two nodes $ i$ and $ j$ (i.e., $( i, j)\in \mathcal{E}$) implies a communication link between the corresponding devices. This graph will be determined by the specific D2D protocol in place among the devices, which may factor in transmit powers, distances, and channel conditions (e.g., see Sec. V of ~\cite{9562522}), which we assume is specified at the wireless layer. Without loss of generality, we assume $\mathcal{G}$ is undirected, i.e.  $( i, j)\in \mathcal{E}$ implies $( j, i)\in \mathcal{E}$, $\forall i, j$, and is static over all time $t$. We further denote the neighbors of device $ i$ with $\mathcal{N}_i =\{ j: ( i, j)\in \mathcal{E} \}$.  
 
 {We consider \textit{cooperation} among the devices~\cite{wang_devicesampling,9311906}
 , where devices push and pull information among themselves over the graph $\mathcal{G}$.
%  , where
%  each device $i$ pushes $\mathcal{I}_{i\rightarrow j}^{t}$ to and pulls $\mathcal{I}_{j \rightarrow i}^{t}$ from each of neighboring devices $j\in\mathcal{N}_i$.  
 In case of explicit information sharing, the pushed/pulled information is a subset of the raw datapoints at the transmitter, while for implicit information exchange, it is a set of embeddings generated at the transmitter}.
 
%  \ali{ by a \textit{selection algorithm} $Q(j,\mathcal{D}_{i\rightarrow j}^{\mathsf{Reserve}},\bm{\phi}_G^{t}) = \mathcal{I}_{j \rightarrow i}^t$ detailed in Sec.~\ref{subsec:B}.}

%  During data transfer between a pair of nodes, the receiving device $i$ \textit{requests} remote datapoints from device $j,~ j \in \mathcal{N}_i$. This process is further detailed in Sec.~\ref{subsec:B}.

% Also, the transmitting  device $j$ shares a fixed number of datapoints with $i$ at every data exchange session. Thus, each device $ i$ obtains remote datapoints when receiving, which are added to the set of datapoints to be used as negatives for training, and shares local datapoints with remote devices $j$ when transmitting. 
% Note that only remote \textit{datapoints} are exchanged exclusively, and not the local \textit{embeddings}. 

% \ali{The following paragraphs are super high level! Let us change the logic! First, describe the loss function per device, then describe the global loss function (all in math!) and then describe how the devices aim to minimize their local loss (gradient descent iterations in math!) and then describe how the global aggregations are performed!}

% \ali{Also, we should not talk about the particular dataset here! Untill we reach the simulation section everythig should be generic!}

\begin{figure}
    \centering
       \hspace{-3.5mm} \includegraphics[width=1.0\columnwidth]{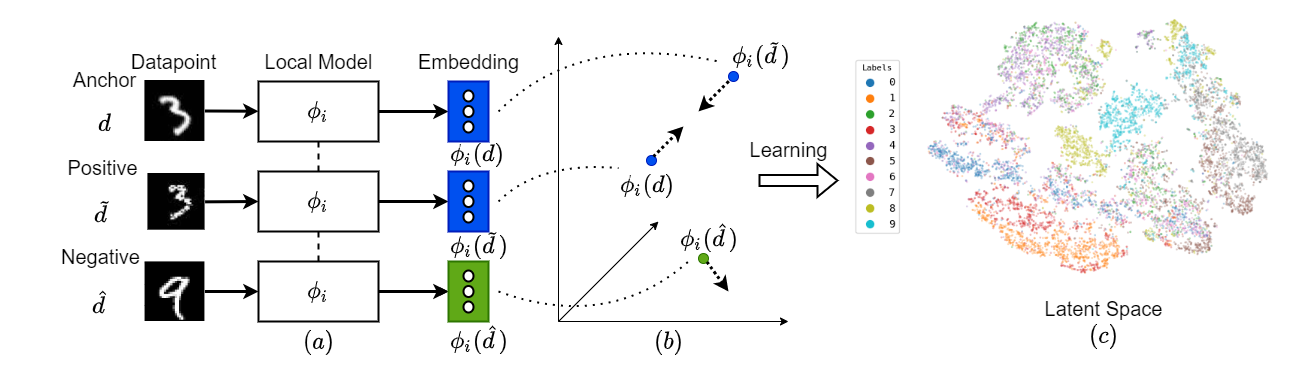}
    \vspace{-2mm}
    \caption{(a) A datapoint (\textit{anchor}), its augmentation (\textit{positive}) and a distinct datapoint (\textit{negative}), are passed through models to obtain embeddings (b). Training maximizes distance between anchor and negative, while minimizing distance between anchor and positive (c).}
    \label{fig:triplet_loss_pipe}
    %%\vspace{-8mm}
\end{figure}

\vspace{-.5mm}
\subsection{Unsupervised FL Formulation}\label{sec:ML}
We consider an unsupervised learning task whose output is a set of \textit{embeddings} of datapoints (i.e., projections of datapoints onto a latent space). We adopt a contrastive learning (CL) framework given its popularity in centralized ML 
%\cite{1640964}, 
\cite{pmlr-v119-chen20j}
. CL aims to learn embeddings by minimizing the \textit{distance} between similar datapoints while maximizing it between dissimilar datapoints. 
% In this paper, we focus on one of the well-known loss functions named \textit{triplet loss} \cite{Dong_2018_ECCV}.
Given an anchor datapoint $d$; a similar datapoint (i.e., a positive) is obtained by applying a randomly sampled augmentation function (e.g., image transformations, Gaussian blurs and/or noise) to it \cite{pmlr-v119-chen20j}
%\cite{moco}
. Any datapoint distinct from the anchor is considered a dissimilar datapoint (i.e., a negative). 

CL aims to learn embeddings using the \textit{triplet loss} function~\cite{Dong_2018_ECCV}.
Formally, given an embedding model $\bm{\phi}$, margin $m$, anchor $d$, augmented anchor (positive) $\tilde{d} = {F}(d)$ and distinct datapoint (negative) $\hat{d}$, {triplet loss} $L$ is defined for  triplet $\{d,\tilde{d},\hat{d}\}$ as 
\vspace{-4mm}

\begin{equation}
    \label{eqn : triplet_loss}
    L_{\bm{\phi}}(d,\tilde{d},\hat{d}) \hspace{-.6mm}=\hspace{-.6mm} \max \hspace{-.7mm}\Bigl\{\hspace{-.1mm}0,\hspace{-.5mm} ||\bm{\phi}(d)-\bm{\phi}(\tilde{d})||_2^2 \hspace{-.9mm}-\hspace{-.9mm} ||\bm{\phi}(d)-\bm{\phi}(\hat{d})||_2^2 \hspace{-.7mm}+\hspace{-.7mm} m\hspace{-.1mm}\Bigr\}\hspace{-.2mm},
    \hspace{-.7mm}
\end{equation}
where ${F}$ is a random augmentation function selected from a set of predefined augmentation functions $\mathcal{F}$ (i.e., $F \in \mathcal{F}$). Minimizing triplet loss promotes a latent space in which similar datapoints are closer to one another while dissimilar ones are further away in latent space by at least a margin of $m$ as illustrated in
Fig \ref{fig:triplet_loss_pipe}.
% , where $d,\tilde{d},\hat{d}$ are the anchor, positive and negative, respectively.

%a non-negative quantity, which takes a non-zero value if the distance between the anchor and the negative exceeds the distance between the anchor and the positive by a value smaller than a margin $m$. An illustration of the triplet loss is given in Fig \ref{fig:triplet_loss}, where $d,\tilde{d},\hat{d}$ are the anchor, positive and negative, respectively.

% Each device $ i$ receives a set of datapoints from remote devices as $\tilde{\mathcal{D}_i} = \{Q(i,j) \forall j \neq i\}$. 

In our distributed ML setting, we define the goal of unsupervised FL as identifying a global model $\bm{\phi}_G^\star$ such that 
\vspace{-1.5mm}
\begin{equation}\label{eq:mainProb}
\vspace{-.7mm}
    \bm{\phi}_G^\star = \min_{\bm{\phi}\in\mathbb{R}^p}~ \sum_{d \in \mathcal{D}} \sum_{F \in \mathcal{F}} \sum_{\hat{d} \in \mathcal{D}, \hat{d}\neq d} L_{\bm{\phi}}(d,{F}(d),\hat{d})\vspace{-1mm},
\end{equation}
where $\mathcal{D} = \bigcup_{i\in\mathcal{C}} \mathcal{D}_i$ represents the global dataset. The  optimal \textit{global latent space} (e.g., subplot (c) in Fig.~\ref{fig:triplet_loss_pipe}) is the one in which anchors and their positive samples are closer to each other while being further away from negative samples across the \textit{global dataset} in the latent space. 

In the federated setting, the local datasets $\mathcal{D}_i$ will not be independent and identically distributed (non-i.i.d.). In order to speed up convergence in the presence of non-i.i.d., \textit{alignment} between \textit{local} latent spaces during training is necessary. 
We propose to accelerate this alignment across devices by smart data (i.e., explicit information) or embedding (i.e., implicit information) transfers. Intuitively, when the data across the devices is homogeneous (i.e., i.i.d.), the latent spaces learned locally will be aligned, under which the global model training will exhibit a fast convergence. Thus, given non-i.i.d. data distributions, we aim to select and share a set of \textit{important} datapoints or embeddings across the devices, that result in the closest alignment of their local models.

\vspace{-.5mm}
\section{Cooperative Federated
\label{sec:coop_fed}
Unsupervised Contrastive Learning ({\tt CF-CL})}
\noindent

In this section, we develop our unsupervised FL methodology {\tt CF-CL}. We first give an overview of the training and information exchange processes of {\tt CF-CL} in Sec.~\ref{sec:train}. We then detail our cooperative data transfer mechanism, i.e., explicit data exchange in Sec.~\ref{subsec:B}. Finally, we detail our embedding exchange methodology, i.e., implicit data exchange in Sec.~\ref{subsec:C}.

\vspace{-.7mm}
%\naji{In an unsupervised setting better alignment of representations over a distributed network speeds up the training process. We are achieving this by resource efficient data transfer}
\subsection{Training and Information Exchange Overview}\label{sec:train}
\vspace{-.3mm}

%To solve~\eqref{eq:mainProb}, we propose {\tt CF-CL} which conducts  distributed ML through a sequence of global model aggregations indexed by $\gamma \in \mathbb{Z}^+$, such that local models, defined in Sec.~\ref{sec:net}, are aggregated at time-steps $t \in \{\gamma T_a\}_{\gamma \in \mathbb{Z}^+}$, where $T_a$ is the aggregation interval. The system is trained for a total of $T$ time-steps, in each time-step, each device $i\in\mathcal{C}$ conducts one mini-batch SGD iteration.
%across devices conducted through a sequence of global model aggregations indexed by $\gamma \in \mathbb{Z}^+$, such that local models, defined in Sec.~\ref{sec:net}, are aggregated at time-steps $t \in \{\gamma T_a\}_{\gamma \in \mathbb{Z}^+}$, where $T_a$ is the aggregation interval. The system is trained for a total of $T$ time-steps, where in each time-step, each device $i\in\mathcal{C}$ conducts one mini-batch SGD iteration.

In {\tt CF-CL},~\eqref{eq:mainProb} is solved through a sequence of global model aggregations indexed by $\gamma \in \mathbb{Z}^+$. Let $T_a \in \mathbb{Z}^{+}$ denote the aggregation interval length in time-steps, so that the aggregations occur at times $t \in \{\gamma T_a\}_{\gamma \in \mathbb{Z}^+}$. At each time-step $t = 1,...,T$, each device $i\in\mathcal{C}$ conducts one mini-batch stochastic gradient descent (SGD) iteration.

Operationally, the information exchange process in {\tt CF-CL} is a combination of
\begin{enumerate}
    \item A \textit{Push} of information from each transmitting device $i\in\mathcal{C}$ to its neighbors $j\in\mathcal{N}_i$. The pushed set is denoted by $\mathcal{D}_{i\rightarrow j}^{\mathsf{Reserve}}$ in the case of explicit information exchange
and $\mathcal{Z}_{i\rightarrow j}^{\mathsf{Reserve}}$ in the case of implicit information exchange.The pushed information is used by device $j$ for determining information importance across its own neighbors.

\item This is followed by a {periodic} \textit{Pull} of information. Indexed by $\tau \in \mathbb{Z}^+$, pulls occur at time-steps $t \in \{\tau T_p\}_{\tau \in \mathbb{Z}^+}$, where $T_p$ is the information pull period. At each $t=\tau T_p$, 
%At each data pull round each $\tau$, at time-step $t=\tau T_p$, 
each receiving device $i$ requests pulling $n^t_{j \rightarrow i}$ information units (i.e., datapoints or embeddings) from device $j \in \mathcal{N}_i$. The resulting pulled information is denoted by $\mathcal{D}_{j \rightarrow i}^{t}$ in case of explicit information exchange and $\mathcal{Z}_{j \rightarrow i}^{t}$ in case of implicit information exchange. 

\end{enumerate}

%\cgb{Edited} 
Periodic push-pull of information enables {\tt CF-CL} to select information that is most important for local model training based on the most recent model parameters. As we observe in (\ref{eqn : triplet_loss}), there is zero loss and thus no change in the model parameters when the difference between the anchor-negative and anchor-positive embedding distances is more than the margin $m$, where the anchor, positive and negative embeddings generated by a model are a function of the current model parameters $\bm{\phi}$. As a consequence, importance calculations based on stale model parameters can lead to the exchange of information that does not produce any meaningful change in the local models, and thus is not an efficient use of communication resources.

%\blue{We emphasize that our work focuses on data transfer in wireless systems which pose crucial resource constraints. In such scenarios, depending on the size of the dataset, sharing the full dataset with neighbours may be infeasible. Additionally, doing so may cause unnecessary exchange of data that is not beneficial to the training of the local model. Literature in the domain of unsupervised learning shows that contrastive learning models benefit more from negative samples which are closer to the anchor sample [14; 15]. This is reflected in the formulation of contrastive loss in Eq. (1), wherein there is no loss and thus no change in the model parameters when the difference between the anchor-negative distance and anchor-positive distance is more than the margin m. Thus, sharing the complete dataset at the beginning would involve sharing a large number of such redundant negatives, both in terms of similarity as well as in terms of not producing any meaningful change in the receiver model. This results in unnecessary utilization of computation and communication resources on transmitting samples that do not contribute to the learning task. Also, the utility of a negative sample is determined by the model parameters and hence, changes over the course of training, thus necessitating regular exchanges of a small amount of data.}

In practice, the number of information units pulled $\{n^t_{j \rightarrow i}\}_{i\in\mathcal{C},j\in\mathcal{N}_i}$ are subject to a limited exchange budget, e.g, based on wireless resource availability. Similar to the D2D graph $\mathcal{G}$, we assume the values of $n^t_{j \rightarrow i}$ are fixed by the wireless layer, and focus on smart information selection built on top of this. Additionally, we assume that each device has a buffer of limited size to store pulled information units.  Hence, before pulling information at time $t = \tau T_p$, it will purge any information pulled in the previous iterations $\tau'$, where $\tau' T_p < \tau T_p$.\footnote{Our method also applies to settings where devices have unlimited buffer sizes and accumulate the pulled information units.}.

The pseudocode of the overall CF-CL procedure is given in Alg. ~\ref{alg:flde}. The push-pull algorithms will be detailed in the following subsections. In doing so, one of our main contributions will be developing the information sampling functions referenced in lines $12$ and $15$ of Alg. ~\ref{alg:flde}. The function for explicit information exchange will sample the most important datapoints for local model training (Alg. ~\ref{algo:explicit_sampling}), while for implicit information exchange it will promote a transfer of embeddings leading to the most expected impact on local training (Alg. ~\ref{algo:implicit_sampling}). 

\vspace{-1mm}
% \ali{Say what is comming next!}\satya{added}

% \satya{Add description of Push-Pull Mechanism - Added to next section}

 \begin{algorithm}[h]
    \caption{{\tt CF-CL} Procedure at each Device $i\in\mathcal{C}$}
    \label{alg:flde}
       {\small
     \textbf{Input:} $i, T_a,T_p, \alpha,$ $\mathcal{D}_i$, $\{n^t_{j \rightarrow i}\}_{j\in\mathcal{N}_i}$
     
     Device $i$ receives the initial global model from the server $\bm{\phi}_G^{0}$
     
     Device $i$ performs K-means clustering on $\mathcal{D}_i$ with $K=K^{\mathsf{Reserve}}_{i \rightarrow j}$  \label{state:2}
     
     Device $i$ samples reserve data $\mathcal{D}^{\mathsf{Reserve}}_{i \rightarrow j}$ by choosing $K^{\mathsf{Reserve}}_{i \rightarrow j}$ datapoints closest to centroids.
     
     \If{{Explicit Information Sharing}}
         {Device $i$ pushes reserve data $\mathcal{D}^{\mathsf{Reserve}}_{i \rightarrow j}$ given by~\eqref{eq:approx_reserve} to each of its neighboring devices $j, ~ j \in \mathcal{N}_i$}\label{state:4}
        % \State Add received datapoints $\mathcal{D}^t_{j \rightarrow i}$ to local dataset
        % \State Update local model $\bm{\phi}_i^t$ according to (\ref{eq:local_update}) \label{state:update_local_model_exp}
    \For{$t=1 $ \textrm{to} $ T$}{
        \If{$t = \tau T_p, \tau  \in \mathbb{Z}^+$}{
            \For{$ j \in \mathcal{N}_i$}{
                 Device $i$ requests pulling $n_{j \rightarrow i}^t$ units of information from device $j$ \label{state:8}
                 \If{{Explicit Information Sharing}}
                  {Device $j$  transmits datapoints to device $i$ using Alg.~\ref{algo:explicit_sampling}}
        \If{Implicit Information Sharing}
        {Device $i$ pushes the embeddings of reserved datapoint $\mathcal{Z}^{\mathsf{Reserve}}_{i \rightarrow j}$ given by \eqref{eq:im_get_reserve_embeddings} to device $j$
         {Device $j$  transmits embeddings to device $i$ using Alg.~\ref{algo:implicit_sampling}}}
         }}
        \If{{Explicit Information Sharing}}
                   {Device $i$ updates its local model $\bm{\phi}_i^t$ according to (\ref{eq:local_update}) using the triplet loss definition in~\eqref{eqn : triplet_loss}}
        \If{Implicit Information Sharing}
        {Device $i$ updates its local model $\bm{\phi}_i^t$ according to (\ref{eq:local_update}) using the regularized triplet loss definition in~\eqref{eq:triplet_loss_reg}}
        %   Device $i$ updates its local model $\bm{\phi}_i^t$ according to (\ref{eq:local_update}) in case of explicit information sharing and uses $L_{\bm{\phi},i,t}(d,\tilde{d},\hat{d})$ given by (\ref{eq:triplet_loss_reg}) in case of implicit information sharing \label{state:update_local_model_exp}.
          \If{$t = \gamma T_a, \gamma \in \mathbb{Z}^+$}
            {Device $i$ sends its local model to the server, which updates global model $\bm{\phi}_G^t$ according to (\ref{eq:aggregation})}}}
\end{algorithm}

% which converges to model $\Tilde{\phi_G}$ such that, 
% \begin{equation}
%     \mathcal{L}_{\Tilde{\phi_G}}(d_i,\tilde{d_i},d_j) < \mathcal{L}_{\hat{\phi_G}}(d_i,\tilde{d_i},d_j) 
% \end{equation}

% Where $\hat{\phi_G}$ is a converged model obtained when selection algorithm $Q(i,j)$ is based on known heuristics.

\subsection{Smart Push-Pull of Explicit Information}\label{subsec:B}
Here we develop our methodology for settings where D2D data sharing is permissible, i.e., in ML applications where privacy is not a major concern.

\subsubsection{Model Training} 
% The reserved datapoints will be later used to determine the important datapoints $\mathcal{D}_{j \rightarrow i}^{t}$ to be pulled by device $i$ from device $j$. $\mathcal{D}_{j \rightarrow i}^{t}$ is periodically selected 
% based on a sampling function $Q$ detailed in~Sec.~\ref{subsec:pulldata}. 

Consider the time-steps $t$ between data pulls $\tau$ and $\tau+1$, i.e., $ \tau T_p\leq t < (\tau+1) T_p$. The data stored at each device will be $\mathcal{D}_i^{t}=\mathcal{D}_i^{\tau T_p}\triangleq \mathcal{D}_i \cup \tilde{\mathcal{D}}_i^{\tau T_p}$ 
, i.e., its initial datapoints ${\mathcal{D}_i}$ and those pulled from neighboring devices $\tilde{\mathcal{D}}^{\tau T_p}_i=\cup_{j\in\mathcal{N}_i} \mathcal{D}^{\tau T_p}_{j \rightarrow i}$. Over the course of the period, device $i$ aims to minimize its triplet loss via solving a local version of (\ref{eq:mainProb}) as follows:
\vspace{-1mm}
% \ali{There is a big confusion here. Why the index of data point is the same as device? (they are both $i$)} \satya{Added datapoint indices to notations along with clilent indices. Each datapoint is associated with a local device dataset and an index.}
% \ali{we need to index the received data points by $\tau$ later since it changes from one global aggregation to another}
% \satya{Would it be useful to define the set of exchanged datapoints as $\tilde{\mathcal{D}}_i^{\tau}$ in that case?} \ali{please see my comment above}
\begin{equation}
\label{eqn:local_loss}
   \bm{\phi}^\star_i = \min_{\bm{\phi}} 
   \sum_{d \in \mathcal{D}_i^t} \sum_{F \in \mathcal{F}} \sum_{\hat{d} \in \mathcal{D}_i^t,~\hat{d}\neq d}
   L_{\bm{\phi}}(d,{F}(d),\hat{d}).
   \vspace{-1mm}
\end{equation}
% Considering (\ref{eqn:local_loss}),
% we have incorporated the data transferred across the nodes exclusively in $\hat{d}$ (i.e., the data transferred across the nodes are only  used as negative samples).
% In our setting the data transfered across the nodes, which are sampled from $\mathcal{D}_i^{\tau T_p} = \mathcal{D}_i \cup \tilde{\mathcal{D}}_i^{\tau T_p}$,  will only be used as negative samples. 
% This sampling configuration is utilized due to the unsupervised setting, where positives are simply augmentations of anchors. Thus, negatives provide the most information for learning an aligned embedding space across devices. Further discussions on the sampling configuration is provided in Sec.~\ref{subsec:B}.

%We only transfer negatives, which are sampled from $\mathcal{D}_i^{\tau T_p} = \mathcal{D}_i \cup \tilde{\mathcal{D}}_i^{\tau T_p}$. Also, anchors $d$ and positives  $\tilde{d}$ are sampled exclusively from the local dataset $\mathcal{D}_i$. We suggest this sampling configuration considering that in our unsupervised setting, where positives are simply augmentations of anchors, negatives provide the most information for learning an aligned embedding space across devices. Further discussions on the sampling configuration is provided in Sec.~\ref{subsec:B}.

% \ali{change the notations!}
 
\begin{algorithmic}[1]
    \begin{algorithm}
    \caption{Explicit Information Pull by Device $i$ from Device   $j\in\mathcal{N}_i$} 
    \label{algo:explicit_sampling}
    {\small
     \Function{Q}{({$j,\mathcal{D}_{i \rightarrow j}^{\mathsf{Reserve}},\bm{\phi}_G^{\gamma T_a},n^t_{j \rightarrow i}$})}
     \State Transmitter $j$ approximates its local dataset as $\mathcal{D}_j^{t,\mathsf{Approx}}$ according to (\ref{eq:approx_local})\label{state:approx_local}
    
    \State \label{macro} Transmitter $j$ performs Kmeans++ on $\mathcal{D}_{j \rightarrow i}^{\mathsf{Reserve}}$ and $\mathcal{D}^{t, \mathsf{Approx}}_j$ \Comment{\textit{Macro Importance}}
    
    \State \label{macro2} Transmitter $j$ obtains sampling probability $[ P^{t,\mathsf{Macro}}_{j \rightarrow i} ({\ell})]_{\ell\in\mathcal{L}^{t}_{j \rightarrow i}}$ via \eqref{eq:ProbBasic}
    % performs K-Means on embeddings of $\mathcal{D}_{i\rightarrow j}^{\mathsf{Reserve}}$ and $\mathcal{D}^{t, \mathsf{Approx}}_j$ to obtain $K$ clusters of embeddings $\{\bm{\phi}_G^{t}(\hat{d})\}_k$ (\eqref{eq:ProbBasic})

    \State \label{micro}Transmitter $j$ calculates importance of each datapoint $\hat{d}$ in cluster $\ell \in\mathcal{L}^{t}_{j \rightarrow i}$ by \eqref{eqn:imp_sampling} \Comment{\textit{Micro Importance}}
    % $\mathcal{D}_j^{t,\mathsf{Approx}}$ as $\beta(\hat{d})$ using $\mathcal{D}_{i \rightarrow j}^{\mathsf{Reserve}}$ as anchors by (\ref{eq:exp_loss}).
    
    \State Transmitter $j$ samples ${n^t_{j \rightarrow i}}$ datapoints to obtain set $\mathcal{D}^t_{j \rightarrow i}$ according to $P^t_{j\rightarrow i}$ from  \eqref{final}    \Comment{\textit{Data sampling}}
    %\Statex \label{exp_exchange} If Explicit Information Sharing; $\mathcal{I}^t_{j \rightarrow i} = \mathcal{D}^t_{j \rightarrow i}$\\
    %\Statex \label{imp_exchange} If Implicit Information Sharing; $\mathcal{I}^t_{j \rightarrow i} = \phi_G^{\gamma T_a}(d) ~\forall~ d \in \mathcal{D}^t_{j \rightarrow i}$\\

    \State Device $j$ transmits  $\mathcal{D}^t_{j \rightarrow i}$ to device $i$ \Comment{\textit{Data Transfer}}
    \EndFunction
% \end{algorithmic}
}
\end{algorithm}
\end{algorithmic}

To solve~\eqref{eqn:local_loss}, each device undergoes local model updates via SGD.
% In particular, at each global aggregation round $\tau$ letting $t\in\mathbb{Z}^+$ denote the index of SGD iteration. 
% At each global aggregation round $\gamma$, i.e., 
At time $t$, given local model $\bm{\phi}_i^{t}$ and a mini-batch of triplets $\mathcal{B}_i^{t}  = \big\{(d,\tilde{d},\hat{d}): {d} \in \mathcal{D}^t_i, \tilde{d} =  {F}(d), {F} \in \mathcal{F} , \hat{d}\in \mathcal{D}^t_i\big\}$ sampled from its local dataset $\mathcal{D}^t_i$, device $i$ updates its local model as
\vspace{-.8mm}
\begin{equation}
    \label{eq:local_update}
    \bm{\phi}_i^{t+1} = \bm{\phi}_i^{t} -\alpha_t  \sum_{(d,\tilde{d},\hat{d}) \in \mathcal{B}_i^t}{\nabla_{\bm{\phi}_i^{t}}L_{{\bm{\phi}_i^{t}}}(d,\tilde{d},\hat{d})}\big/{|\mathcal{B}_i^{t}|},
    \vspace{-.8mm}
\end{equation}
where $\alpha_t$ is the learning rate at time $t$.
%  This process is described in lines $12$ to $15$ in Algorithm \ref{alg:flde}. \ali{This is coming out of the blue. It is still not clear that what is Alg 1. What happened to the first 12 lines? Instead, At the end of this subsection, put a paragraph explaining all the lines and then say efficient data sampling is explained in the next subsection.}\satya{Changed.}
% \ali{we need to add a figure for $T_p$ and $T_a$.. I will talk about that with you in the meeting. Also, the time is not concretized (will talk in the meeting).}
% \ali{non-iid and local bias...still this is not added!}\satya{TODO : Add notes on local bias and non i.i.d ness - DONE}
% Note that without data exchange, $\hat{d}$ would be sampled from local dataset $\mathcal{D}_i$. If all datapoints in minibatch $\mathcal{B}^t$ for triplet loss are sampled from local dataset, this introduces a bias towards the local data at device $i$. In a realistic federated setting, this local data may be non i.i.d. Thus, if local models are biased towards local data in such a scenario, rate of convergence of model is significantly reduced. We illustrate this phenomenon in the next section.

To solve~\eqref{eq:mainProb}, the server aggregates the local models $\{\phi_i^{t+1}\}_{i \in \mathcal{C}}$. Aggregation $\gamma$ occurs at $t = \gamma T_a$, and generates a global model $\bm{\phi}^{{t}}_G$. We consider an aggregation scheme proportional to the average cardinality of local datapoints at each device since the last aggregation round $(\gamma-1)$. Formally, letting $D_i^{(\gamma-1:\gamma)} =  \sum_{t \in \{(\gamma-1) T_a+1\cdot\cdot\cdot\gamma T_a\}}{D_i^t}/{T_a}$, $\forall i$, the aggregation is:
\begin{equation}
\vspace{-.8mm}
    \label{eq:aggregation}
   \hspace{-3mm} \bm{\phi}^{t}_G = \frac{1}{\sum\limits_{i\in\mathcal{C}}D_i^{(\gamma-1:\gamma)}}\sum_{i\in\mathcal{C}} \bm{\phi}^{t}_i{D_i^{(\gamma-1:\gamma)}},~t = \gamma T_a,\hspace{-.7mm}~\gamma\in\mathbb{Z}^+  \hspace{-.7mm}.  \hspace{-3mm}
    \vspace{-.8mm}
\end{equation}
 Global model $\bm{\phi}_G^t$ is then broadcast across all devices $i \in \mathcal{C}$, synchronizing all local models, i.e., $\bm{\phi}_i^{t}\leftarrow\bm{\phi}^{t}_G,$ when $t = \gamma T_a$. This model is then used for the next local training rounds as in~\eqref{eq:local_update}.

% \ali{This is the most important section of the paper! Using the notations that we have defined above, describe your data transfer method in math combined with explanation and intuition. Then add a psudo-code to make your method crystal clear!} 

% \ali{All the temperature and stuff should also be described here! In general, the reader should not be introduced to a new method unless it is a baseline in the simulations section.}

% \ali{Start off this section by saying the big idea behind your method. Are you trying to avoid the bias? Are you trying to push the gradeint somewhere? Are you trying to allign the local grads with each other? Are you trying to alling the local gradients with the global gradient?  let us have a big idea here and then you can move to the details}

%  thus propose a smart data push-pull strategy that can eliminate the impact of non-iid data and improve the training convergence speed.
\subsubsection{Smart Push of Explicit Information}\label{sec:PushKmeans}
Explicit information exchange aims to share datapoints that 
%best capture the modes of the local data distributions, identifying important datapoints that 
best contribute to cross-device embedding alignment. First, each device $i$ pushes a set of \textit{representative} datapoints (called reserve data) to each of its neighbors $j$ as $\mathcal{D}_{i\rightarrow j}^{\mathsf{Reserve}}$ at $j$, i.e,  
% can improve the performance dramatically, specifically in scenarios where not a lot of data can be pushed.
\vspace{-1mm}
    \begin{align}
        \label{eq:approx_reserve}
        \mathcal{D}^{\mathsf{Reserve}}_{i \rightarrow j} = \{d: d \sim \mathcal{D}_i\},~ |\mathcal{D}^{\mathsf{Reserve}}_{i \rightarrow j}| = K^{\mathsf{Reserve}}_{i \rightarrow j},~j\in\mathcal{N}_i,
    \end{align}
    %   \vspace{-5mm} 
where $K^{\mathsf{Reserve}}_{i \rightarrow j}$ is the number of reserve datapoints at $j$ from $i$. To select explicit information to exchange, we use {K-means++} \cite{kmeans} clustering, with $K=K^{\mathsf{Reserve}}_{i \rightarrow j}$ at each device $i$ and include the centers of clusters/centroids in
$\mathcal{D}^{\mathsf{Reserve}}_{i \rightarrow j}$, $j\in\mathcal{N}_i$.
Typically, $K^{\mathsf{Reserve}}_{i \rightarrow j}$ is constant across neighbours $j$, so K-means++ is executed only once. This increases  performance significantly compared to random data sampling, especially if $K^{\mathsf{Reserve}}_{i \rightarrow j}$ is small, as we will see  in Sec.~\ref{experiments}.

% \textbf{\textit{Pulling Reserve Datapoints : }}
% % Before initiating training of local models, we allow for provisioning of a subset of datapoints from each remote device at the transmitter, referred to as reserve datapoints $\mathcal{D}^{\mathsf{Reserve}}_{j}$.
% Transmitter $i$ is allowed to pull $\mathcal{D}^{\mathsf{Reserve}}_{j \rightarrow i} ~ \forall ~ j \in \mathcal{N}_i$ such that it is available locally exclusively for importance calculation.

% \textbf{\textit{Utilization of Remote Datapoints as Negatives:}} We use datapoints sampled from remote devices as negatives. As in our unsupervised setting positives are generated as augmentations of anchor, while negatives are any other distinct datapoint.\satya{A possible rationale here could be that if we were to use remote data as anchors, that would increase the calculation complexity as we will have to to account for importance of combinations of remote datapoints and augmentations. It would benefit performance, but it would also increase computational complexity by a factor of $2|\mathcal{F}|$}
%The pushed data is then used as a reserve set based on which the importance of samples are calculated. Important data points are defined as the ones that contribute more to alignment.

\subsubsection{Smart Pull of Explicit Information} \label{subsec:pulldata}
% Our data selection strategy $Q$ is summarized in Alg.~\ref{algo:explicit_sampling} (incorporated into {\tt{CF-CL}} in Alg. \ref{alg:flde})
We now describe the strategy for explicit information selection, which we denote by $Q$. The summary of this process is given in in Alg.~\ref{algo:explicit_sampling}.
$Q$ identifies locally important datapoints to be pulled by each device $i\in\mathcal{C}$ from device $ j\in\mathcal{N}_i$. At each global aggregation time-step $t=\gamma T_a$, device $j$ approximates its local dataset by uniformly sampling a fixed number $K^{\mathsf{Approx}}_j$ of its local datapoints, i.e,
% At device $i$, we approximate the distribution at device $i$ via 
%constituting the set $\mathcal{D}^{ t,\mathsf{Approx}}_j$. $\mathcal{D}^{t,\mathsf{Approx}}_j$ constitutes the set of candidate datapoints at device $j$ for transmission to neighboring devices. It is obtained as follows
    \vspace{-1mm}
    \begin{align}
        \label{eq:approx_local}
       \hspace{-2mm} \mathcal{D}^{ t, \mathsf{Approx}}_j = \{d: \hspace{-.5mm} d \sim \mathcal{D}^{t}_j\},\hspace{-.5mm}~|\mathcal{D}^{ t,\mathsf{Approx}}_j| \hspace{-.5mm}=\hspace{-.5mm} K^{\mathsf{Approx}}_j,\hspace{-.5mm}~t=\gamma T_a. \hspace{-2mm}
    \end{align}
$\mathcal{D}^{t,\mathsf{Approx}}_j$ constitutes the set of candidate datapoints at device $j$ for transmission to neighboring devices. Uniformly sampling a subset of $\mathcal{D}^{t}$ improves the efficiency of {\tt{CF-CL}} for large local datasets while providing an unbiased estimate of the local data distribution at device $j$. 

At each information pull instance $\tau$, which occurs between two global aggregation rounds $\gamma$ and $\gamma +1$ (i.e., $\gamma T_a \leq \tau T_p < (\gamma+1) T_a $), the explicit information pull by device $i$ from device $j$ is denoted by
 $ \mathcal{D}^{\tau T_p}_{j \rightarrow i}= Q(j,\mathcal{D}_{i\rightarrow j }^{\mathsf{Reserve}},\bm{\phi}_G^{\gamma T_a},n^t_{j \rightarrow i})\subseteq \mathcal{D}^{ t, \mathsf{Approx}}_j$.
We aim to design the selection strategy $Q$ to promote faster convergence of global models $\bm{\phi}_G^{t}$ to the optimal model $\bm{\phi}^{\star}_G$ by sampling and pulling information that is \textit{important} in that, it accelerates local model convergence while avoiding local model bias.
The global model $\bm{\phi}_G^{\gamma T_a}$  is used in $Q$ to determine the most effective datapoints from device $j$ to minimize device $i$'s bias to its local dataset. 

 Formally, to perform the data pull between each pair of devices $(i,j)$, we implement a \textit{two-stage probabilistic importance sampling} procedure, consisting  of macro and micro sampling steps.

\textit{(1) Macro Sampling:} In \textit{macro sampling}, the data at device $j$ that is used for estimates pertaining to device $i$, which is $\mathcal{D}_{i\rightarrow j}^{\mathsf{Reserve}}\cup \mathcal{D}^{t, \mathsf{Approx}}_j$, is partitioned into clusters, which are assigned a cluster-level sampling probability. Specifically, device $j$ obtains the embeddings of all datapoints in $\mathcal{D}_{i\rightarrow j}^{\mathsf{Reserve}}$ and $\mathcal{D}^{t, \mathsf{Approx}}_j$ by feeding them through global model $\bm{\phi}_G^{\gamma T_a}$, and performs K-means++ to partition them into clusters of embeddings, which we denote as a set of clusters $\mathcal{L}^{t}_{j\rightarrow i}$.
%  Naturally, the clusters which contain a higher ratio of datapoints from $\mathcal{D}^{t, \mathsf{Approx}}_j$ to $\mathcal{D}_{i\rightarrow j}^{\mathsf{Reserve}}$ contain the datapoints that are less similar to the data distribution at device $i$, constituiting the important datapoints.
    Device $j$ then assigns a sampling probability $P^{t,\mathsf{Macro}}_{j\rightarrow i} ({\ell})$ to each of the  clusters $\ell\in \mathcal{L}^{t}_{j\rightarrow i}$ proportional to the importance of the cluster relative to device $i$; we refer to this cluster level importance as the \textit{macro} sampling probability. Formally, these probabilities are obtained as
     \vspace{-1mm}
     \begin{equation}\label{eq:ProbBasic}
         P^{t,\mathsf{Macro}}_{j \rightarrow i} ({\ell}) =  \frac{X^{t,\mathsf{Macro}}_{j \rightarrow i} ({\ell})}{ \sum_{\ell\in \mathcal{L}^{t}_{j \rightarrow i}} X^{t,\mathsf{Macro}}_{j \rightarrow i} ({\ell})},~t=\tau T_p,
          \vspace{-1mm}
     \end{equation}
     where
      \vspace{-1mm}
     \begin{equation}\label{eq:xX}
       X^{t,\mathsf{Macro}}_{j \rightarrow i} ({\ell}) \triangleq \frac{K^{t,\mathsf{Approx}}_{j \rightarrow i}(\ell)}{{K^{t,\mathsf{Approx}}_{ j}(\ell)+K^{t,\mathsf{Reserve}}_{i\rightarrow j}(\ell)}}.
        \vspace{-1mm}
     \end{equation}
     In~\eqref{eq:xX}, $K^{t,\mathsf{Approx}}_{j \rightarrow i}(\ell)$ is the number of datapoints in $\mathcal{D}^{t, \mathsf{Approx}}_j$ located in cluster $\ell$,  and $K^{t,\mathsf{Reserve}}_{i\rightarrow j}(\ell)$ is the number of datapoints in $\mathcal{D}^{t, \mathsf{Reserve}}_{i\rightarrow j}$ located in cluster $\ell$.\footnote{Note that we have $\sum_{\ell\in \mathcal{L}^{t}_j}K^{t,\mathsf{Approx}}_{j \rightarrow i}(\ell)=K^{\mathsf{Approx}}_j,\forall t$ and  $\sum_{\ell\in \mathcal{L}^{t}_j}K^{t,\mathsf{Reserve}}_{i\rightarrow j}(\ell) = K^{\mathsf{Reserve}}_{i\rightarrow j},\forall t$.}
    Intuitively, using the sampling probability $P^{t,\mathsf{Macro}}_{j \rightarrow i} ({\ell}) $ defined in \eqref{eq:ProbBasic} promotes selection of clusters containing a higher fraction of the datapoints at device $j$, which are representative of $j$'s distribution (through $K_j^{t, Approx}(\ell)$) but dissimilar from device $i$ (through $K_{i \rightarrow j}^{t,Reserve}(\ell)$), and thus selecting clusters which are more likely to improve device $i$'s local distribution. Such diversification of the local distribution at each device $i$ results in the local dataset being closer to the global distribution that FL is aiming to optimize over. The benefit of local dataset diversification has been established analytically in existing works for supervised FL, e.g., \cite{henry_infocom_data_offloading,psl}.
    %\cgb{these refs need to be linked!}.  
    CF-CL builds upon this theoretical foundation, as the push-pull mechanism employed by each device assigns importance to datapoints/embeddings at their neighbors according to the dissimilarity from their local distribution.
    
    \textit{(2) Micro Sampling:}  In \textit{micro sampling} the probability of sampling individual datapoints are calculated to assign a data-level importance separate from the cluster assignments via \eqref{eq:ProbBasic}. We assign a probability $P^{t,\mathsf{Micro}}_{j \rightarrow i}  ({\hat{d}})$ to data point $\hat{d}$ according to the expected loss when it is used as a negative relative to datapoints in $\mathcal{D}_{i\rightarrow j}^{\mathsf{Reserve}}$ used as anchors:
% In particular, at data pull instance $t=\gamma T_a$,
% to calculate the approximate \textit{importance} of local datapoint $\hat{d}\in \mathcal{D}^{ t, \mathsf{Approx}}_i$, transmitter $i$ calculates the
% triplet loss for $\hat{d}$ against $d \in \mathcal{D}^{\mathsf{Reserve}}_{i\rightarrow  j}$ based on global model $\bm{\phi}_G^t$,  corresponding to 
% the expected loss $\mathbb{E}_{d \sim \mathcal{D}^{\mathsf{Approx}}_i}\left[L_{\bm{\phi}^t_i}({d},{\tilde{d}},\hat{d})\right]$ 
% of local model $\bm{\phi}_i^t$ for datapoint $\hat{d}$ against local anchors ${d} \in \mathcal{D}^{\mathsf{Approx}}_i$ and local positives $\tilde{d} \in \tilde{\mathcal{D}}^{\mathsf{Approx}}_i$
% defined as 
\vspace{-2.5mm}

%\hspace{-6mm} \beta(\hat{d}) = \mathbb{E}_{d \sim \mathcal{D}^{\mathsf{Reserve}}_{i\rightarrow  j}}\hspace{-.2mm}\left[L_{\bm{\phi}^t_G}({d},{\tilde{d}},\hat{d})\right]  \hspace{-.2mm}  \hspace{-.2mm}=\hspace{-.2mm} \frac{\sum_{d \in \mathcal{D}^{\mathsf{Reserve}}_{i\rightarrow  j}} L_{\bm{\phi}^t_G}(d,\tilde{d},\hat{d})}{|\mathcal{D}^{\mathsf{Reserve}}_{i\rightarrow  j}|}, \tilde{d} \hspace{-.2mm}=\hspace{-.2mm} F(d). \hspace{-4mm} 

\begin{equation}
    \label{eq:exp_loss}
   \hspace{-2mm}\mathbb{E}_{\hspace{-.2mm}d \sim \mathcal{D}^{\mathsf{Reserve}}_{i\rightarrow  j}} \hspace{-.2mm}\hspace{-1mm} \left[\hspace{-.2mm}\hspace{-.3mm}L_{\hspace{-.3mm}\bm{\phi}^{\gamma T_a}_G}(\hspace{-.2mm}d,\hspace{-.2mm}F(d),\hspace{-.2mm}\hat{d})\hspace{-.3mm}\hspace{-.2mm}\hspace{-.1mm}\right] \hspace{-.2mm}\hspace{-1mm}=\hspace{-1mm} \frac{\hspace{-.6mm}\sum_{d \in \mathcal{D}^{\mathsf{Reserve}}_{i\rightarrow  j}}\hspace{-.2mm} L_{\hspace{-.2mm}\bm{\phi}^{\gamma T_a}_G}\hspace{-.5mm}(\hspace{-.2mm}d,\hspace{-.2mm}\tilde{d},\hspace{-.2mm}\hat{d})}{K^{\mathsf{Reserve}}_{i\rightarrow  j}}\hspace{-.2mm}.
\end{equation}
\vspace{-2.5mm}

\noindent 
% Given $\mathcal{\hat{D}}^{\mathsf{Approx}}_j$ for all neighbors $j \in \mathcal{N}_i$, we define the buffer of received datapoints at device $i$ as $\mathcal{D}_i^{\mathsf{Arrival}} = \bigcup_{j \in\mathcal{N}_i} \mathcal{\hat{D}}^{\mathsf{Approx}}_j$. 
The probability of selection of datapoint $\hat{d}\in\ell$ is then computed as 
\vspace{-3mm}

\begin{equation}
    \label{eqn:imp_sampling}
  \hspace{-3mm} P^{t,\mathsf{Micro}}_{j \rightarrow i}  ({\hat{d}}) = 
    \frac{\exp\left(\lambda^t \cdot \mathbb{E}_{d \sim \mathcal{D}^{\mathsf{Reserve}}_{i\rightarrow  j}}\hspace{-1mm}\left[L_{\bm{\phi}^{\gamma T_a}_G}(d,\tilde{d},\hat{d})\right]\right)}{  \hspace{-2mm} \sum_{\hat{d}' \in  \ell} \exp\left(\lambda^t \cdot \mathbb{E}_{d \sim \mathcal{D}^{\mathsf{Reserve}}_{i\rightarrow  j}} \hspace{-1mm} \left[L_{\bm{\phi}^{\gamma T_a}_G}(d,\tilde{d},\hat{d}')\right]\right)}.  \hspace{-3mm} 
\end{equation}
\vspace{-1.5mm}

 In \eqref{eqn:imp_sampling}, $\lambda^t$ is the \textit{selection temperature}, a user-defined hyperparameter which is used
% As the loss observed on local models decreases with training, the difference between losses of any two datapoints shrinks. Thus, without a specified selection temperature, the selection scheme would devolve to a uniform selection mechanism when the model is close to convergence. 
  to make our selection algorithm robust against the loss values becoming more homogeneous. Considering~\eqref{eqn:imp_sampling}, our selection algorithm aims to improve the model training performance by prioritizing the transmission of datapoints which produce a higher loss at the receiver (measured via their loss over $\mathcal{D}_{i\rightarrow j }^{\mathsf{Reserve}}$). Intuitively, the choice of $\lambda^t$ defines the degree of greediness with which the algorithm selects important datapoints.

Combining \textit{macro sampling} and \textit{micro sampling}, when device $i$ pulls $n_{j \rightarrow i}^t$ datapoints from $j$, they are selected from device $j$'s candidate samples $\mathcal{D}^{t,\mathsf{Approx}}_j$. The probability of sampling of each datapoint $\hat{d} \in \mathcal{D}_j^{\mathsf{Approx}}$ given by
\vspace{-1mm}
\begin{equation}\label{final}
    P^{t}_{j \rightarrow i}  ({\hat{d}})= P^{t,\mathsf{Micro}}_{j \rightarrow i}  ({\hat{d}})\times P^{t,\mathsf{Macro}}_{j \rightarrow i} ({\ell}_{\hat{d}}), \hat{d} \in \mathcal{D}_j^{\mathsf{Approx}},
\end{equation}
Where $\ell_{\hat{d}}$ is the cluster that $\hat{d}$ is assigned to via the K-means++ procedure that generates $\mathcal{L}_{j \rightarrow i}$.

\subsection{Smart Push-Pull of Implicit Information and Triplet Loss Regularization}\label{subsec:C}

In Sec. \ref{subsec:B}, we allowed for explicit information exchange in the form of data. We now consider the exchange of embeddings or implicit information, which is smaller in size and also more desirable in privacy sensitive applications. These benefits come at the cost of performance compromises compared to data exchange, as we will see experimentally in Sec.~\ref{experiments}. 
% These representations cannot be inverted to obtain datapoints, and hence privacy is preserved in this form of implicit information exchange. 

 When implicit information is exchanged between devices, the local ML training and global model aggregations of {\tt CF-CL} are similar to those under explicit information exchange and follow~\eqref{eq:local_update} and \eqref{eq:aggregation} with two differences: (i) The datapoints used for training at each device are static (i.e., $\mathcal{D}^t_i=\mathcal{D}_i,~\forall t$), since only the embeddings are exchanged, and (ii) a modified definition for the triplet loss function $L_{\bm{\phi}}(d,\tilde{d},\hat{d})$ is used which includes a regularization term to incorporate exchanged implicit information, which is explained in this section.

Similar to explicit push-pull, our method consists of pushing embeddings from each device to its neighboring devices, and are used to identify local embeddings that contribute the most to model performance, followed by pulling important embeddings from the neighboring devices.

\subsubsection{Smart Push of Implicit Information} At each global aggregation time-step $t=\gamma T_a$, $\forall \gamma$, each device first pushes reserve embeddings $\mathcal{Z}_{i \rightarrow j}^{t,\mathsf{Reserve}}$ to its neighboring devices,
providing them with fresh embeddings to use for importance measurements.
To this end, each device $i$ samples a set of representative local datapoints $\mathcal{D}_i^{t,\mathsf{Reserve}}$ according to \eqref{eq:approx_reserve} and obtains their embeddings $\mathcal{Z}_{i \rightarrow j}^{t,\mathsf{Reserve}}$, which are then pushed to device $j$, where\footnote{As compared to explicit push-pull in Sec.~\ref{subsec:B}, where data push is conducted only once at the beginning of training, in implicit push-pull we need to conduct the push procedure after every global aggregation since the embeddings are generated using the current global model.}
\begin{align}
    \label{eq:im_get_reserve_embeddings}
  \hspace{-3mm}  \mathcal{Z}_{i \rightarrow j}^{t,\mathsf{Reserve}} = \left\{z: z = \bm{\phi}_G^{\gamma T_a}(d), d \in \mathcal{D}_i^{t,\mathsf{Reserve}}\right\}\hspace{-.8mm},\hspace{-.5mm}~t=\gamma T_a. \hspace{-3mm}
\end{align}

{\begin{algorithmic}[1] 
    \begin{algorithm}[t!]
    \caption{Implicit Information Pull by Device $i$ from Device   $j\in\mathcal{N}_i$}
    \label{algo:implicit_sampling}
    {\small
% \begin{algorithmic}[1]
    % \State \textbf{Input:} Device $i$, Remote device $j$, selection temperature $\lambda$
    %  \Function{Q}{$j,\mathcal{Z}_{i \rightarrow j}^{\mathsf{Reserve}},\bm{\phi}_G^{\gamma T_a},n^t_{j \rightarrow i}$}\\
    \Function{Q'}{({$j,\mathcal{Z}_{i \rightarrow j}^{\mathsf{Reserve}},\bm{\phi}_G^{\gamma T_a},n^t_{j \rightarrow i}$})}
     \State \label{smp_2} Transmitter $j$ receives reserve embeddings $\mathcal{Z}_{i \rightarrow j}^{t,\mathsf{Reserve}}$ from device $i$ as (\ref{eq:im_get_reserve_embeddings})\label{state:im_get_reserve_embeddings}\\
     \State \label{smp_2} Transmitter $j$ approximates local dataset $\mathcal{D}_j^{t,\mathsf{Approx}}$ as (\ref{eq:approx_local})\label{state:approx_local}\\
     \State \label{im_macro3} Transmitter $j$ obtains local embeddings $\mathcal{Z}_{j}^{t,\mathsf{Approx}}$ as (\ref{eq:im_get_local_embeddings})\\
    \State \label{im_macro2} Transmitter $j$ performs KMeans++ to obtain clusters of local embeddings $\mathcal{H}^t_{j}$ \Comment{\textit{Macro Importance}}\\
    % ~~~~\label{im_macro3} Transmitter $j$ obtains importance of each cluster according to (\ref{eq:im_cluster_importance})\\
    \State \label{im_macro3} Transmitter $j$ obtains probability distribution for clusters according to (\ref{eq:im_cluster_probs}) \\
    % performs K-Means on embeddings of $\mathcal{D}_{i\rightarrow j}^{\mathsf{Reserve}}$ and $\mathcal{D}^{t, \mathsf{Approx}}_j$ to obtain $K$ clusters of embeddings $\{\bm{\phi}_G^{t}(\hat{d})\}_k$ (\eqref{eq:ProbBasic})
    % ~~~~\label{im_micro1} Transmitter $j$ obtains importance of each local embedding according to (\ref{eq:im_emb_importance})\\
    \State \label{im_micro2} Transmitter $j$ obtains probability distribution for embeddings in each cluster according to (\ref{eq:im_emb_probs}) and (\ref{eq:emb_overlap})\\ \Comment{\textit{Micro Importance}}\\
    % $\mathcal{D}_j^{t,\mathsf{Approx}}$ as $\beta(\hat{d})$ using $\mathcal{D}_{i \rightarrow j}^{\mathsf{Reserve}}$ as anchors by (\ref{eq:exp_loss}).
    % \For{${n^t_{j \rightarrow i}}$}{
    % ~~~~\label{smp} Transmitter $j$ selects cluster $\ell$ according to (\ref{eq:im_cluster_probs})  \\
    \State \label{smp} Transmitter $j$ samples ${n^t_{j \rightarrow i}}$ embeddings to obtain set $\mathcal{Z}^t_{j \rightarrow i}$ according to probabilities obtained in (\ref{eq:overall_selection_prob_embed}) \\ \Comment{\textit{Embedding Sampling}}\\
    % }
    \State \label{smp} Transmitter $j$ transmits  $\mathcal{Z}^t_{j \rightarrow i}$ to device $i$ \Comment{\textit{Embedding Transfer}}
    \EndFunction
}
\end{algorithm}
\end{algorithmic}
}

\subsubsection{Smart Pull of Implicit Information} \label{smart_pull}

Similar to data sampling in Sec.~\ref{subsec:B}, our embedding selection method consists of: (i) identifying and sampling clusters of embeddings proportional to their importance; and (ii) the selection of high importance embeddings from the sampled clusters. A detailed description of these processes is given in Sec.~\ref{smart_pull}.

%\cgb{Updated} 
In the case of implicit exchange, local embeddings are a function of the global model $\phi_G^{\gamma T_a}$, and thus change over the course of training. Hence, as with explicit information, it is necessary for the implicit information to be shared regularly over time so that it reflects the current global model.
%First, we would like to highlight that local embeddings are a function of the parameters of the corresponding global model, according to Eqs. (13)\&(14) in the manuscript. Thus, they will change over training rounds, i.e., as $\phi_G^{\gamma T_a}$ updates. Hence, it is necessary for the embeddings to be shared regularly over time so that they reflect the current local model as opposed to sharing all embeddings to each neighbour at the start of training, which will only be representative of the global model at the start of the training process. Embeddings which reflect the current global model are crucial in order to ensure that the exchange process promotes transfers that most strongly improve alignment relative to the present local training process. This also helps prevent overlap between embeddings generated by different devices. 

At each embedding pull instance $\tau$, i.e., $t=\tau T_p$, which occurs between two global aggregation rounds $\gamma$ and $\gamma +1$ (i.e., $\gamma T_a \leq \tau T_p < (\gamma+1) T_a $), the embeddings pulled by device $i$ from device $j$, denoted by set $\mathcal{Z}_{j \rightarrow i}^{t}$, are obtained by execution of probabilistic function $Q'$, where $\mathcal{Z}_{j \rightarrow i}^{t}=Q'(j,\mathcal{Z}_{i \rightarrow j}^{\gamma T_a,\mathsf{Reserve}},\bm{\phi}_G^{\gamma T_a},n^t_{j \rightarrow i})\subseteq \{\bm{\phi}_j^t(d)\}_{d \in \mathcal{D}_j}$. We provide a summary of $Q'$ in Alg.~\ref{algo:implicit_sampling} and detail it below.
After $\mathcal{Z}_{j \rightarrow i}^{t}$ is obtained, device $j$ first selects a set of representative local datapoints $\mathcal{D}_j^{t,\mathsf{Approx}}$ according to~\eqref{eq:approx_local} and generates their embeddings $\mathcal{Z}_{j}^{t,\mathsf{Approx}}$ as follows:
\begin{align}
    \label{eq:im_get_local_embeddings}
    \hspace{-3mm}\mathcal{Z}_{j}^{t,\mathsf{Approx}} = \left\{z: z = \bm{\phi}_G^{\gamma T_a}(d), d \in \mathcal{D}_j^{t,\mathsf{Approx}}\right\},~t=\tau T_p. \hspace{-3mm}
\end{align}

Elements of $\mathcal{Z}_{j}^{t,\mathsf{Approx}}$ are candidate embeddings from which the most important are to be selected and transmitted to the neighboring devices. Note that for each neighboring device $i\in\mathcal{N}_j$, most important embeddings in $\mathcal{Z}_{j}^{t,\mathsf{Approx}}$ are dissimilar since the data distribution at the devices is non-i.i.d. We select the most important embeddings using a \textit{two-stage probabilistic importance sampling} procedure, consisting  of macro and micro sampling steps, similar to Sec.~\ref{subsec:pulldata}.

\textit{(1) Macro Sampling:} partitions $\mathcal{Z}_{j}^{t,\mathsf{Approx}}$ using KMeans++ to $Z^{\mathsf{local}}$ clusters of local embeddings denoted by set $\mathcal{H}^t_{j}$. Similarly, KMeans++ is performed on $\mathcal{Z}_{i \rightarrow j}^{t,\mathsf{Reserve}}$ to obtain $Z^{\mathsf{Reserve}}$ clusters.
    
    For exchange from device $j$ to neighboring device $i$, the sampling probability distribution for each cluster $h$ in $\mathcal{H}^t_j$, is found by calculating the aggregate importance of the embeddings within that cluster using reserve embeddings $\mathcal{Z}_{i \rightarrow j}^{t,\mathsf{Reserve}}$ via the score metric
    % we calculate the importance of $z_{j,l} \in \mathcal{Z}_{j,\ell}^{t,\mathsf{Approx}}$ as follows
    \begin{align}
        \label{eq:im_cluster_importance}
        S(h,\mathcal{Z}_{i \rightarrow j}^{t,\mathsf{Reserve}}) = \frac{\sum_{z \in h} s(z,\mathcal{Z}_{i \rightarrow j}^{t,\mathsf{Reserve}})}{|\mathcal{H}^t_j[h]|},
    \end{align}
    where $|\mathcal{H}^t_j[h]|$ denotes the number of datapoints in cluster $h$.
    In~\eqref{eq:im_cluster_importance}, $s(z,\mathcal{Z}_{i \rightarrow j}^{t,\mathsf{Reserve}})$ is the importance score of embedding $z$ relative to the embeddings in $\mathcal{Z}_{i \rightarrow j}^{t,\mathsf{Reserve}}$ computed as
    %\begin{align}
    %    \label{eq:im_emb_importance}
    %    s(z,\mathcal{Z}_{i \rightarrow j}^{t,\mathsf{Reserve}}) = \sum_{z' \in \mathcal{Z}_{i \rightarrow j}^{t,\mathsf{Reserve}}} \max\left(0,M-\Vert z'-z\Vert^2_{_2}\right),
    %\end{align}
    \begin{align}
        \label{eq:im_emb_importance}
        s(z,\mathcal{Z}_{i \rightarrow j}^{t,\mathsf{Reserve}}) = \max_{z \in h}(0,\Vert z-\mu_h\Vert^2_{_2})\sum_{z' \in \mathcal{Z}_{i \rightarrow j}^{t,\mathsf{Reserve}}} \left(\Vert z'-z\Vert^2_{_2}\right),
    \end{align}
    %where $M$ is a tunable threshold.
    where $\mu_h$ is the centroid of cluster $h$ if embedding $z$ belongs to cluster $h$.

    Intuitively, this formulation ensures that embeddings that are closer to the centroid of the cluster are selected more frequently, and thus the set of selected embeddings is a better representative of its parent cluster. Our importance computation rule in \eqref{eq:im_emb_importance} is tailored to the triplet loss computation rule \eqref{eqn : triplet_loss} and assigns a higher importance score to those embeddings which are closer to the embeddings in $\mathcal{Z}_{i \rightarrow j}^{t,\mathsf{Reserve}}$, since such embeddings constitute the \textit{hard negatives} (i.e., they contribute more to the local training). Using the cluster importance score~\eqref{eq:im_cluster_importance}, we obtain the sampling probability of each cluster $h$ as follows:
    \begin{align}
        \label{eq:im_cluster_probs}
        P^{t,\mathsf{Macro}}_{j \rightarrow i} (h) = \frac{S(h,\mathcal{Z}_{i \rightarrow j}^{t,\mathsf{Reserve}})}{\sum_{h' \in \mathcal{H}_j^t}S(h',\mathcal{Z}_{i \rightarrow j}^{t,\mathsf{Reserve}})}.
    \end{align}

    Transmitting embeddings from clusters which have a significant overlap with $Z^{\mathsf{Reserve}}$ should be avoided, as these clusters have a high probability of containing embeddings similar to $\mathcal{Z}_{i \rightarrow j}^{t,\mathsf{Reserve}}$, which will negatively impact performance if used as regularizers. We calculate the overlap $B(h)$ of local clusters $h$ with centroid $c^{h}$ as follows:
    \begin{align}
        B(h) = \mathsf{PDF}(b(h),\hat{\mu},\hat{\sigma}),
    \end{align}
    \begin{align}
        \label{eq:emb_overlap}
        b(h) = \frac{\frac{\sum_{i \in Z^{\mathsf{Reserve}}}{||c^h - \hat{c}^i||_2^2}}{|Z^{\mathsf{Reserve}}|} - \frac{\sum_{i \in Z^{\mathsf{Local}}}{||c^h - c^i||_2^2}}{|Z^{\mathsf{Local}}|-1}}{\frac{\sum_{i \in Z^{\mathsf{Local}}}{||c^h - c^i||_2^2}}{|Z^{\mathsf{Local}}|-1}},
    \end{align}
where $\mathsf{PDF}$ is the probability density function of a normal distribution. $\hat{\mu},\hat{\sigma}$, the mean and standard deviation of the distribution respectively, are tunable parameters. %We set these to $1$ and $0.5$ respectively for our experiments. 

\begin{figure}[t!]
    \centering
    \includegraphics[width=1.0\columnwidth]{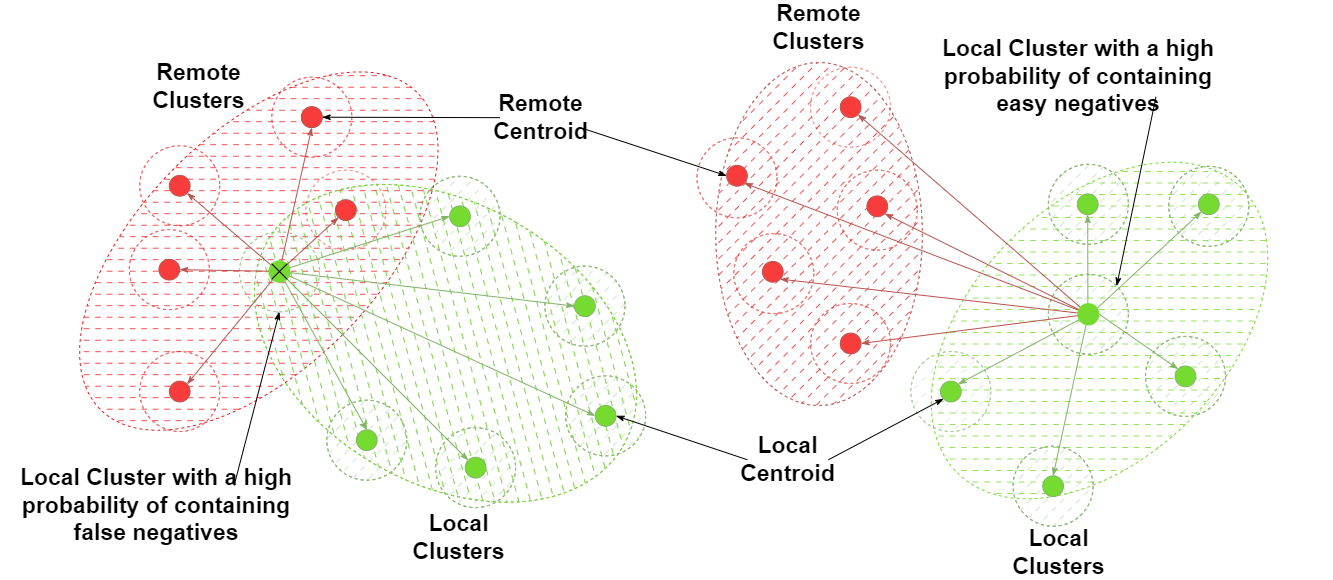}
    \vspace{-2mm}
    \caption{Centroids of remote clusters (green) that are closer to local clusters (red) have a higher chance of being similar to local data (Left), while centroids that are further away are less likely (Right).}
    \label{fig:triplet_loss}
    \vspace{-0.25in}
\end{figure}

$B(h)$ functions as a scaling factor which is large for clusters that are equally close to remote and local clusters, indicating that there is a high likelihood of hard negatives being present in the cluster. Conversely, close proximity to remote clusters relative to local clusters indicate a higher possibility of false negatives and closer proximity to local clusters relative to remote clusters indicate the presence of easy negatives, and hence, the value $B(h)$ for both cases is small, as shown in Fig. \ref{fig:triplet_loss} We modify the probability of cluster $h$ given in \eqref{eq:im_cluster_probs} as follows:
%\begin{align}
%    P^{t,\mathsf{Macro}}_{j \rightarrow i} (h)  =
    %\begin{cases}
     %   P^{t,\mathsf{Macro}}_{j \rightarrow i} (h)   ,& \text{if } B(h) \geq M^{\mathsf{Overlap}}\\
     %   k,                                          & \text{otherwise},k \ll 1
    %\end{cases}
%\end{align}
    \begin{align}
         P^{t,\mathsf{Macro}}_{j \rightarrow i} (h) = B(h)\cdot  P^{t,\mathsf{Macro}}_{j \rightarrow i} (h).
    \end{align}
    
    \textit{(2) Micro Sampling:} In \textit{micro sampling}, once $P^{t,\mathsf{Macro}}_{j \rightarrow i}$ is obtained, the sampling probability of each embedding $z$ belonging to cluster $h$ is calculated as follows:
    \begin{align}
        \label{eq:im_emb_probs}
       P^{t,\mathsf{Micro}}_{j \rightarrow i} (z) = \frac{s(z,\mathcal{Z}_{i \rightarrow j}^{t,\mathsf{Reserve}})}{\sum_{z' \in h}s(z',\mathcal{Z}_{i \rightarrow j}^{t,\mathsf{Reserve}})},~z\in h.
    \end{align}

    Where $s(z,\mathcal{Z}_{i \rightarrow j}^{t,\mathsf{Reserve}})$ is obtained from \eqref{eq:im_emb_importance}.

Combining \textit{macro sampling} and \textit{micro sampling}, at each $t=\tau T_p$, $\forall \tau$, upon pulling $n_{j \rightarrow i}^t$ embeddings by device $i$ from $j$, the embeddings from $\mathcal{D}^{t,\mathsf{Approx}}_j$ are selected to obtain set $\mathcal{Z}_{j \rightarrow i}^{t}$, where $n_{j \rightarrow i}^t=|\mathcal{Z}_{j \rightarrow i}^{t}|$, with each embedding $z$ sampled with probability 
\begin{align}
    \label{eq:overall_selection_prob_embed}
    P^{t}_{j \rightarrow i}  (z) = P^{t,\mathsf{Macro}}_{j \rightarrow i} (h) \times P^{t,\mathsf{Micro}}_{j \rightarrow i} (z),~z\in h,~h\in\mathcal{H}^t_j.
\end{align}
Note that in the above procedure, we obtain $\mathcal{Z}_{j \rightarrow i}^{t}$ only at embedding exchange instances $t=\tau T_p, ~\forall t$. Nevertheless, we will define $\mathcal{Z}_{j \rightarrow i}^{t}$, $1\leq t\leq T$ to facilitate our explanations in the following discussions.
Since the previously received embeddings are kept at each device until receiving new embedding, we define $\mathcal{Z}_{j \rightarrow i}^{t}\triangleq\mathcal{Z}_{j \rightarrow i}^{t}|_{t=\tau T_p}$, where $\tau T_p \leq t <(\tau+1) T_p$. 
%\ali{Where is Fig 3 referred in the main text?}\\
%\ali{Put Fig 3 on the same page as you are referring to it.}

\subsubsection{Integrating Implicit Information into Local ML Training via Triplet Loss Regularization}
As opposed to explicit information exchange, implicit information cannot be directly used in the local ML training as they are generated according to the local model at each transmitting device. Hence, the received device cannot modify them according to its own local model during SGD iterations~\eqref{eq:local_update} since their corresponding input datapoints are not locally available. 

In order to incorporate the information contained in the exchanged embeddings for local training at the receiver, we propose integrating the received embeddings at each device $i$ into its local ML model training by introducing a regularization term into the triplet loss formulation~\eqref{eqn : triplet_loss}. In particular, we redefine the triplet loss at each device $i$ as follows for the duration of local learning $\gamma T_a \leq t < (\gamma+1) T_a $:
\begin{equation}\label{eq:triplet_loss_reg}
     \resizebox{\hsize}{!}{%
     $
 \begin{aligned}
    &L_{\bm{\phi},i,t}(d,\tilde{d},\hat{d}) \hspace{-.6mm}=\underbrace{\hspace{-.6mm} \max \hspace{-.7mm}\bigg[\hspace{-.1mm}0,\hspace{-.5mm} \Vert\bm{\phi}(d)-\bm{\phi}(\tilde{d})\Vert_{_2}^2 \hspace{-.9mm}-\hspace{-.9mm} \Vert\bm{\phi}(d)-\bm{\phi}(\hat{d})\Vert_{_2}^2 \hspace{-.7mm}+\hspace{-.7mm} m\hspace{-.1mm}\bigg]\hspace{-.2mm}}_{\textsf{Contrastive Loss}} 
    +\hspace{-.5mm} \underbrace{W_{t}\hspace{-.5mm}\sum_{j\in\mathcal{N}_i}\sum_{z \in \mathcal{Z}^{t}_{j \rightarrow i}}\hspace{-2mm} \max  \hspace{-.7mm}\bigg[\hspace{-.1mm}0,\hspace{-.5mm} \Vert\bm{\phi}(d)-\bm{\phi}(\tilde{d})\Vert_{_2}^2 \hspace{-.9mm}-\hspace{-.9mm} \Vert\bm{\phi}(d)-{z}\Vert_{_2}^2 \hspace{-.7mm}+\hspace{-.7mm} m^{\mathsf{Reg}}\hspace{-.1mm}\bigg]}_{\textsf{Regularization Term}},
    % + \sum_{i \in k_{rx}} ||\phi(d) - \bar{z_i}||^2_2.
\end{aligned}$
}
\end{equation}

In (\ref{eq:triplet_loss_reg}), the regularization margin $m^{\mathsf{Reg}}$ is different from the triplet loss margin $m$, and adapts to the current model as follows. For local clusters $h \in Z^{\mathsf{Local}}$ with radii $r^h$ we define $m^{\mathsf{Reg}}$ as
\begin{align}
    m^{\mathsf{Reg}} = k \cdot \frac{\sum_{h \in Z^{\mathsf{Local}}} r^h} {|Z^{\mathsf{Local}}|}.
\end{align}

The regularization margin $m^{\mathsf{Reg}}$ ensures that for any anchor embedding in cluster $h$, all remote embeddings that lie within $h$ are considered hard negatives. The scaling factor $k$ indicates the volume of latent space that the model considers when looking for hard negatives. The objective of the regularization term is to incorporate the hard negatives identified via~\eqref{eq:im_emb_importance} into the local ML training. Since the received embeddings at each device $i$, i.e.,  $\cup_{j\in\mathcal{N}_i}\mathcal{Z}^{t}_{j \rightarrow i}$, are computed at the instance of global model reception, they become obsolete as the local ML model training proceeds during the period $\gamma T_a \leq t < (\gamma+1) T_a$. Hence, the regularization coefficient $W_{t}$ is introduced in~\eqref{eq:triplet_loss_reg}, which is used to put a higher weight on the received embeddings while they are fresh (i.e., at the beginning of local ML training) and gradually diminishes as the local model changes and the embeddings become obsolete.
% avoid regions of the latent space occupied by the embeddings of other devices. Not doing so results in regions of overlap between devices, resulting in models that perform poorly. This information is encapsulated by the received embeddings. Hence, minimizing the regularizing term results in maximizing the distance between local embeddings and the regions of latent space occupied by embeddings of the remote devices.
% We note that as opposed to explicit information, implicit information is ephemeral, in that after every local gradient descent iteration $t \rightarrow t+1$, remote embeddings of datapoints $\phi^{t+1}_j(d)$ will differ from the ones exchanged, which is $\phi^{t}_j(d)$.  
Subsequently, to capture the effect of the staleness of received embeddings, we propose the following policy for the regularization coefficient:
\begin{align}
    \label{eq:im_reg_weight}
    W_{t} =  \lambda \cdot\left(\exp \left(\frac{-t~ \mathbf{mode}~T_a}{T_a-1}\right) + \exp\left( \frac{t}{T} -\varrho\zeta_t\right)\right).
\end{align}

The first term in $W_t$ forms a sawtooth shape which takes its maximum value at the time-step of global aggregation $t=\gamma T_a$, $\gamma\in\mathbb{Z}^+$. This term naturally puts a higher weight on the regularization when the received embeddings are fresh. The second term reflects the fact that as the models are trained, model embedding spaces become less transient, and the effect of staleness reduces.

It is worth mentioning that regularization terms have been previously explored in FL research via the well known work~\cite{fedprox} and its subsequent literature in a different context.
Implementations such as \cite{fedprox} consider proximal regularization terms imposed on the model parameters (i.e., neural network parameters) to enhance the convergence of FL over non-i.i.d data. 
The focus of our work and in turn our regularization term exploits the information exchange in unsupervised FL, which is fundamentally different from previous literature focusing on model parameter weights. Hence, the regularization term in \cite{fedprox} can be further mounted on top of our technique.

\section{Numerical Experiments}
\label{experiments}
% \noindent In this section, we first describe the simulation setup (Sec.~\ref{sim:set}) and then provide the numerical results (Sec.~\ref{sim:res}).
% \subsection{Simulation Setup}\label{sim:set}

% \satya{To be Organized}
% \satya{I will be adding in a paragraph about the kinds of simulations we are doing and how we are measuring performance}
% \ali{This is still unorganized! First, start with the description of data sets. Describe the hyperparmeters youhave tuned for each. Then move to Description of the figures. For each figure, first describe what is the goal of the figure and what kind of simualtion is performed in it. And then say the main takeaway of the figure: For example, ``Fig .. depicts the impact of the number of data points shared between the devices on the convergence perforamnce. As can be seen, ...... This suggests that..... "}

\subsection{Simulation Setup}
In our numerical experiments, we use the Fashion MNIST dataset \cite{fmnist}, the USPS Handwritten Digits dataset \cite{uspsdataset} and the Street View House Numbers (SVHN) dataset \cite{svhndataset}. The Fashion MNIST dataset and SVHN dataset consist of $60$K images with $10$ classes and the USPS dataset consists of $7291$ images with $10$ classes.
We consider a network of $|\mathcal{C}|=10$ devices and emulate non-i.i.d. data across devices as discussed below. 

For Fashion MNIST, each device is allocated $6$K datapoints from only three out of $10$ classes. For this dataset, we use the Alexnet architecture \cite{alexnet} with an output size of $16$. The Adam optimizer is used with an initial learning rate of $10^{-4}$ and devices' models are trained for $T = 2000$ local SGD iterations. Data augmentation consists of random resized crops, random horizontal flips, and Gaussian blurs \cite{shorten2019survey}. Unless otherwise stated, we set $K^{\mathsf{Push}}_{i \rightarrow j}=20$, and $K^{\mathsf{Approx}}_j=100$, and local K-means employs $20$ clusters.

For USPS, each device is allocated $\sim 730$ datapoints from only three of $10$ classes. For this dataset, we use a CNN with a single convolutional layer with $8$ kernels, each of size $3 \times 3$, followed by $3$ linear layers of sizes $1024,256$ and $16$. The Adam optimizer is used with an initial learning rate of $10^{-3}$ and models of devices are trained for $T = 1500$ local SGD iterations. Data augmentation consists of random rotations and random perspective transformations. We set $K^{\mathsf{Push}}_{i \rightarrow j}=10$, and $K^{\mathsf{Approx}}_j=100$, and local K-means employs $10$ clusters.

For SVHN, each device is allocated $6$K datapoints from only three out of $10$ classes. For this dataset, we use the Resnet-18 architecture \cite{resnet} with an output size of $256$. The Adam optimizer is used with an initial learning rate of $10^{-4}$ and devices' models are trained for $T = 4000$ local SGD iterations. Data augmentation consists of random resized crops, random horizontal flips, and Gaussian blurs. %\cite{shorten2019survey}
Unless otherwise stated, we set $K^{\mathsf{Push}}_{i \rightarrow j}=25$, and $K^{\mathsf{Approx}}_j=200$, and local K-means employs $25$ clusters.

We conduct simulations on a server with 48GB Tesla-P100 GPU with 128GB RAM. All hyperparameters are identical for implicit and explicit data exchange algorithms.

% Selection temperature is chosen such that it increases linearly as $\lambda^t=6(t/T) + 4$.
% \satya{We use K-Means with $4$ clusters}

% We add random color jitter and random grayscale transformations for CIFAR10.

% \textbf{Evaluation Details.}
% \label{sec:eval_details}
% We observe the cumulative loss of the global model $\phi_G$ over all datapoints $d$ in $\hat{\mathcal{D}} = \bigcup \mathcal{D}_i ~ \forall ~ i$ when used as anchors. We sample negatives $\tilde{d}$ stochastically from $\hat{\mathcal{D}}i$ and positives $\hat{d} = F(d)$, where $F$ is sampled from $\mathcal{F}$. The loss at epoch $t$ is calculated on global model $\phi_G^t$ as $\mathcal{L}_{\phi_G^t}(d,\tilde{d},\hat{d}) \forall d \in \hat{\mathcal{D}}$.
To obtain the accuracy of predictions, we adopt the linear evaluation \cite{pmlr-v119-chen20j}, and use $\bm{\phi}^t_G$, $\forall t$, to train a linear layer $\bm{\theta}$ in a supervised manner on top of $\bm{\phi}^t_G$ to perform a classification at the server.  The linear layer is trained over $1000$ SGD iterations.
As mentioned in Sec.~\ref{intro}, smart data transfer has not been studied in the context of unsupervised federated learning, and literature \cite{wang_devicesampling},\cite{zhao2018federated},\cite{furl} have only considered uniform data transfer across the network. Thus, we compare the performance of {\tt CF-CL} against four baselines: (i) \textit{uniform sampling}, where data points transferred are sampled uniformly at random from the local datasets; 
%\cgb{updated} 
(ii) \textit{bulk sampling}, where importance sampling and data sharing is done only at the beginning of training, and the amount of information shared is equivalent to what CF-CL would share over the course of training; (iii) \textit{K-Means exchange}, where the information to be shared is selected by the transmitter using K-Means clustering (i.e., in place of lines 3 to 5 of Algorithm 2 and lines 2 to 7 of Algorithm 3), with information that is closest to the centroid of the clusters selected to be shared; and (iv) classic federated learning (FedAvg), which does not conduct any data transfer across devices.

The communication graph $\mathcal{G}$ is assumed to be random geometric graph (RGG), which is a common model used for wireless peer-to-peer networks. We follow the same procedure as in~\cite{9705093} to create RGG with average node degree $7$. For Fashion MNIST, devices conduct  $T_a=25$ local SGD iterations and exchange data after $T_p=25$ iterations, and for USPS, $T_a=10$ and $T_p=25$.

% \subsection{Results and Discussions}\label{sim:res}
% Below, we study different characteristics of {\tt CF-CL}.

%  We observe the results of varying the maximum number of datapoints that can be transferred between each pair of nodes $(j,i)$, $|\tilde{\mathcal{D}}_{j,i}^{\tau T_p}| \in \{5,10\}$, number of local samples that approximate local dataset $K^{\mathsf{Push}}$, number of remote samples that approximate remote dataset $K^{\mathsf{Approx}}$, and exchange interval $T_p$. We also observe the effect of delaying the start of data exchange and stopping the data exchange earlier before the training ends.

% We present our experimental results in Fig.~\ref{fig:3}, which describes the effect of our remote datapoint importance sampling algorithm on two metrics of interest, training loss and classification accuracy, which are calculated as described in the evaluation details of section \ref{sec:eval_details}. 
% \ali{Start each of these enumitems with ``In Fig. XX, we study/investigate/depict/ .... . Then explain the intuition behind it. I am revising enumitem 1. Please do the rest and then I will take a pass.}

% \begin{enumerate}[leftmargin=5mm]

\begin{figure*}[t]
        \begin{subfigure}[b]{0.3\textwidth}
		\centering
		\includegraphics[height=4.5cm]{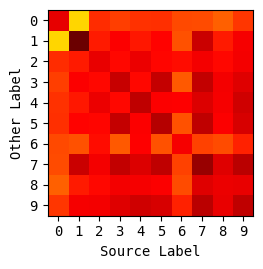}
            \caption{No Information Exchange}
            \label{fig:f7_left}
        \end{subfigure}
        \hfill
        \begin{subfigure}[b]{0.3\textwidth}
		\centering
		\includegraphics[height=4.5cm]{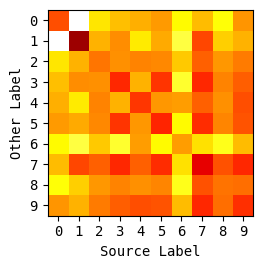}
            \caption{Explicit Information Exchange}
            \label{fig:f7_center}
        \end{subfigure}
        \hfill
        \begin{subfigure}[b]{0.3\textwidth}
		\centering
		\includegraphics[height=4.5cm]{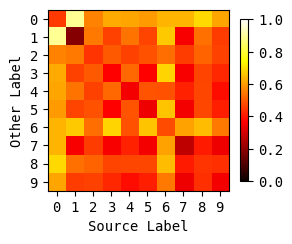}
            \caption{Implicit Information Exchange}
            \label{fig:f7_right}
        \end{subfigure}
		\caption{Distance between embeddings of pairwise combinations of labels. Information exchange results in dissimilar embeddings being further apart from each other.}\label{fig:emb_distances}
  %\vspace{-0.35in}
\end{figure*}

\begin{figure*}[h!]
    \begin{subfigure}[b]{0.3\textwidth}
        \centering
        \includegraphics[height=4cm]{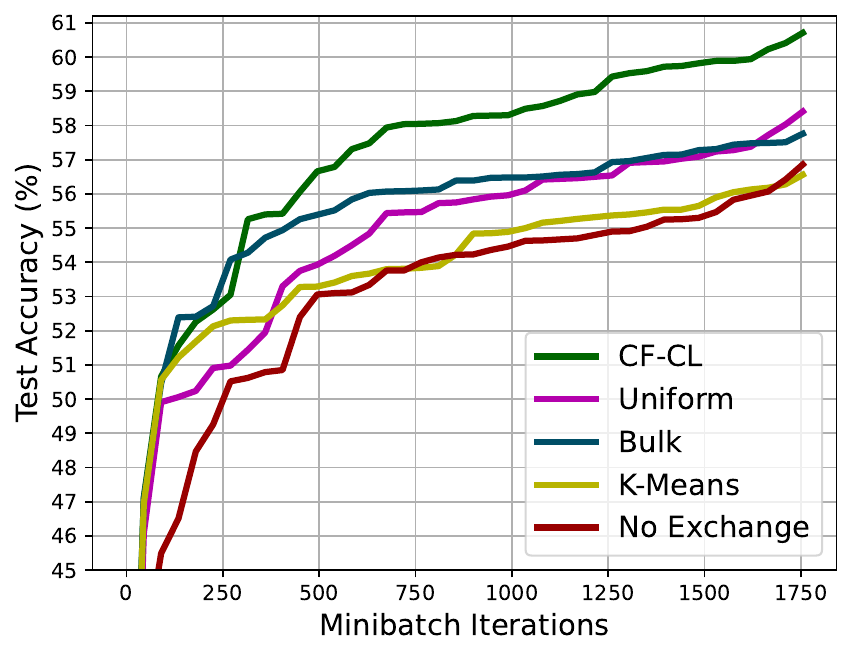}
            \caption{FMNIST Explicit Test Accuracies}
            \label{fig:f5_fmnist_accs_exp}
        \end{subfigure}
    \hfill
    \begin{subfigure}[b]{0.3\textwidth}
        \centering
        \includegraphics[height=4cm]{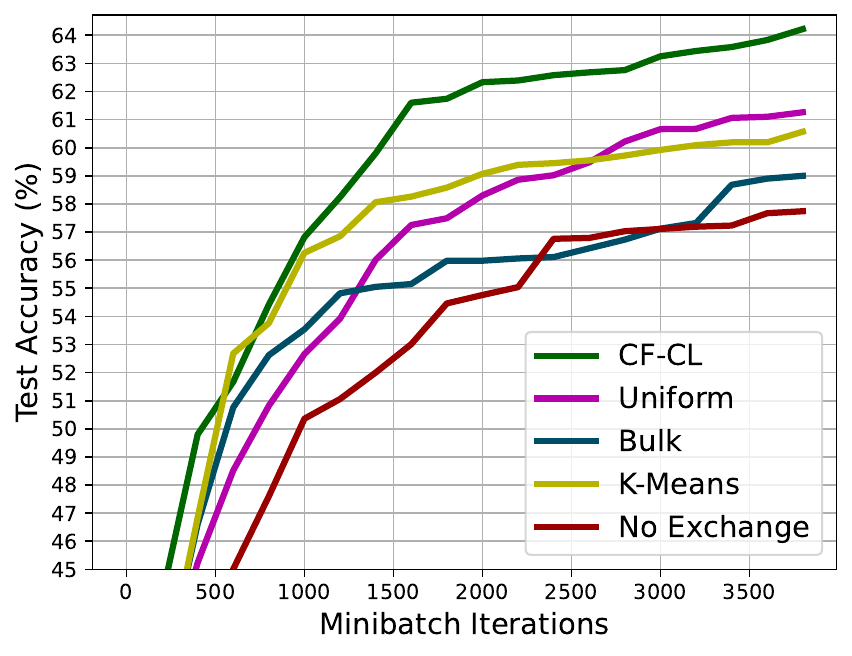}
            \caption{SVHN Explicit Test Accuracies}
            \label{fig:f5_svhn_accs_exp}
        \end{subfigure}
    \hfill
    \begin{subfigure}[b]{0.3\textwidth}
        \centering
        \includegraphics[height=4cm]{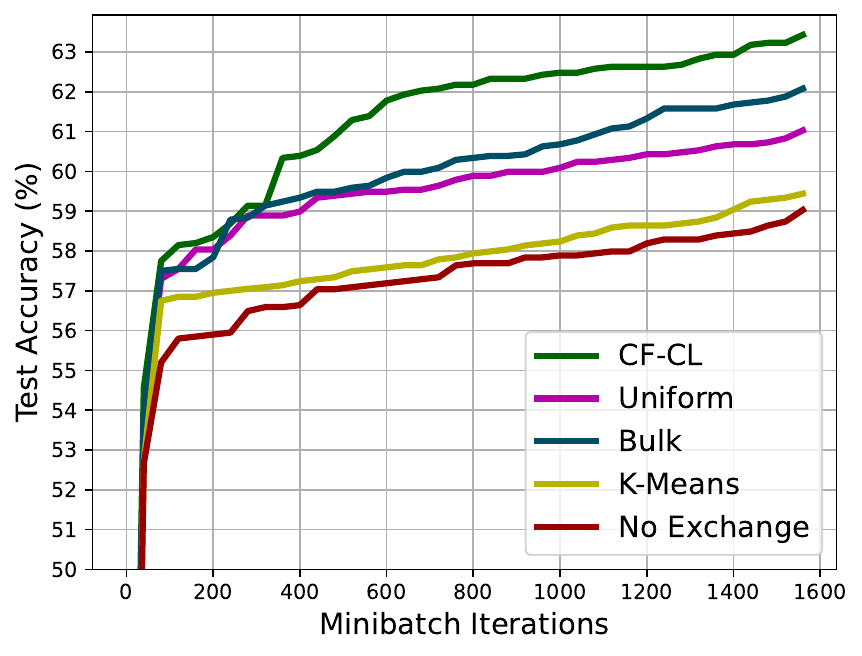}
            \caption{USPS Explicit Test Accuracies}
            \label{fig:f5_usps_accs_exp}
        \end{subfigure}
    \hfill

    \begin{subfigure}[b]{0.3\textwidth}
        \centering
        \includegraphics[height=4cm]{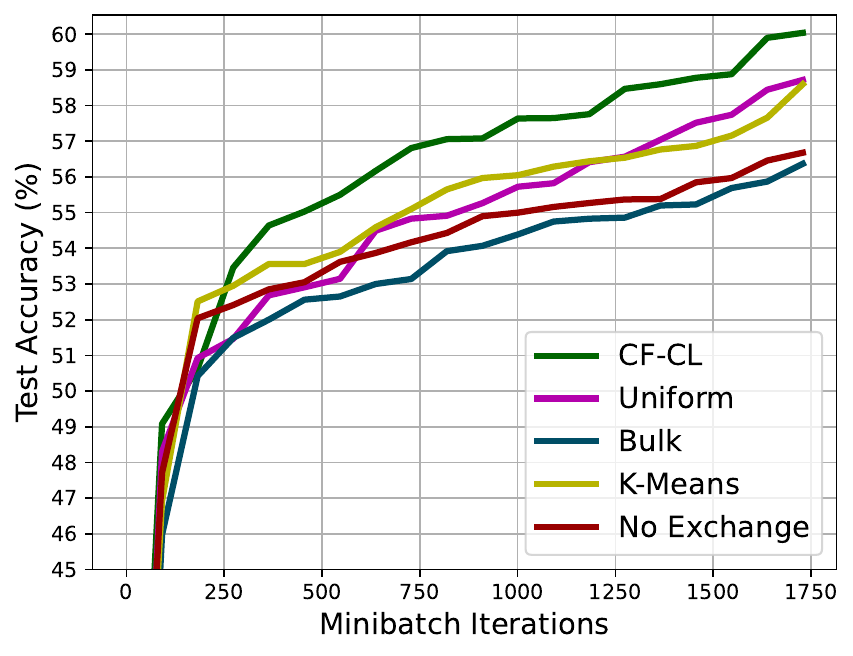}
            \caption{FMNIST Implicit Test Accuracies}
            \label{fig:f5_fmnist_accs_imp}
        \end{subfigure}
    \hfill
    \begin{subfigure}[b]{0.3\textwidth}
        \centering
        \includegraphics[height=4cm]{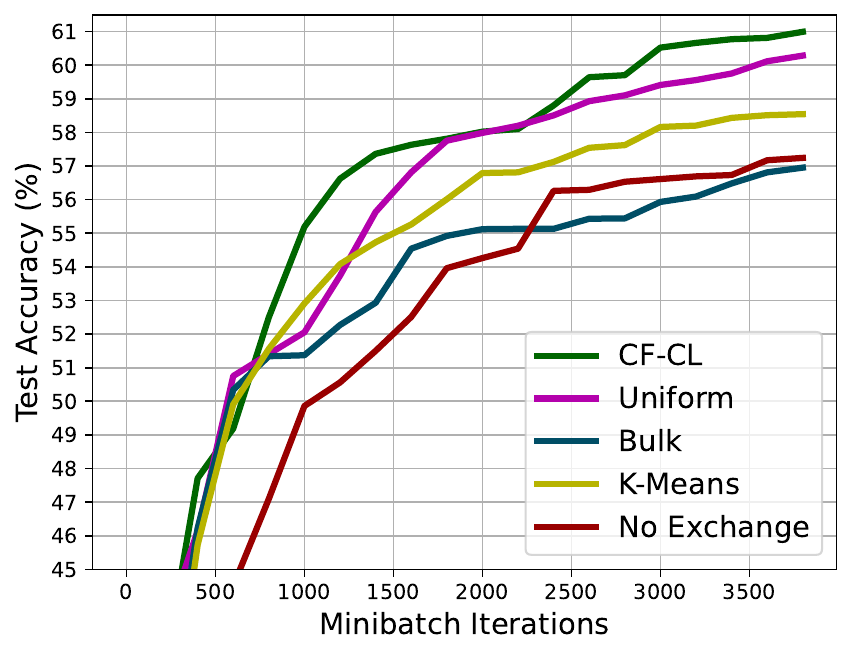}
            \caption{SVHN Implicit Test Accuracies}
            \label{fig:f5_svhn_accs_imp}
        \end{subfigure}
    \hfill
    \begin{subfigure}[b]{0.3\textwidth}
        \centering
        \includegraphics[height=4cm]{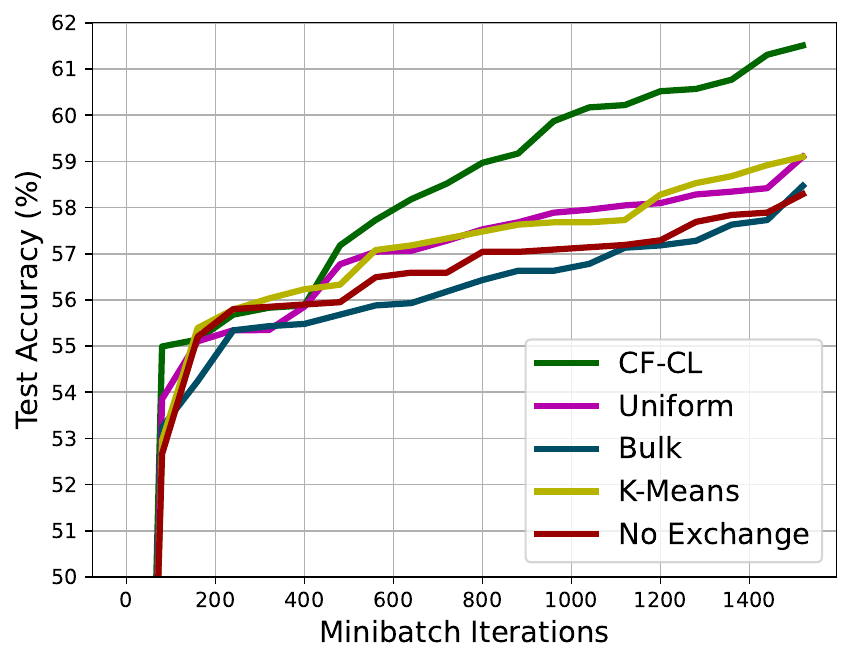}
            \caption{USPS Implicit Test Accuracies}
            \label{fig:f5_usps_accs_imp}
        \end{subfigure}
    \hfill
    \caption{Training performance comparison of {\tt CF-CL} against baselines over training iterations, for each dataset and information exchange regime. We see that {\tt{CF-CL}} has superior performance in all cases, validating the benefit of its information sharing mechanism for local model alignment.}\label{fig:accs}
%\vspace{-8mm}
\end{figure*}

%\fi

%\iffalse
\begin{figure*}[h!]
    
    \begin{subfigure}[b]{0.3\textwidth}
    \centering
    \includegraphics[height=4cm]{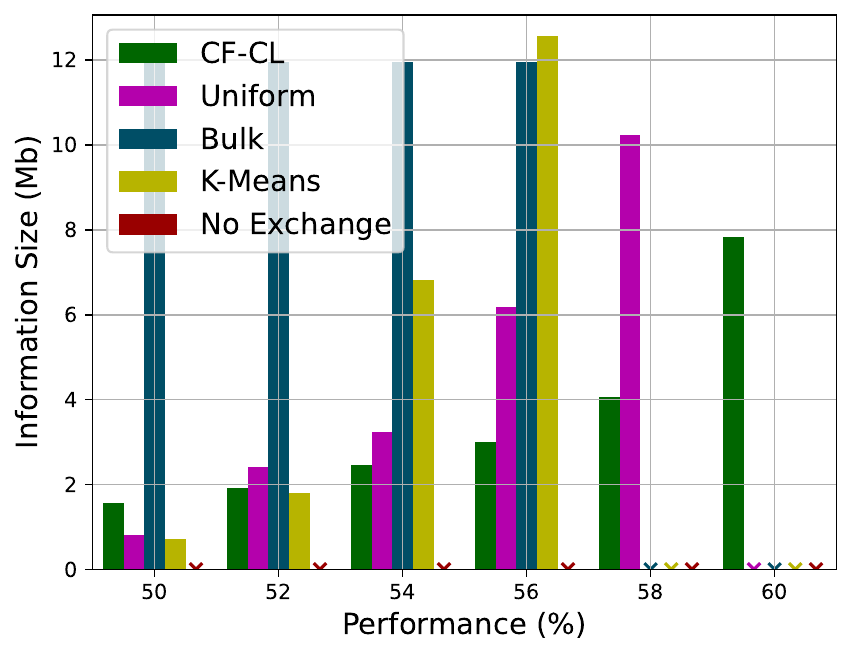}
        \caption{FMNIST Explicit Info. Overheads}
        \label{fig:f5_fmnist_info_exp}
    \end{subfigure}
    \hfill
    \begin{subfigure}[b]{0.3\textwidth}
    \centering
    \includegraphics[height=4cm]{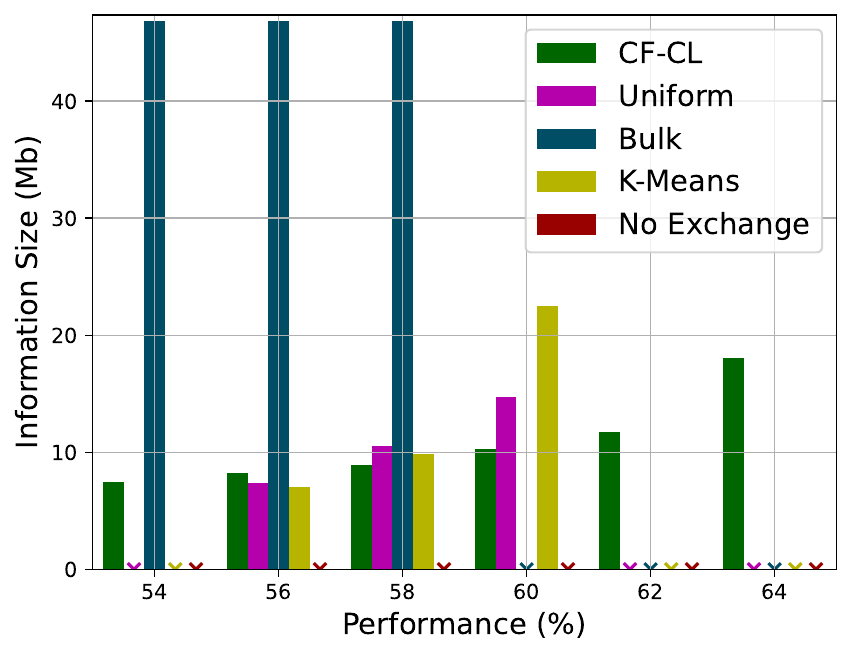}
        \caption{SVHN Explicit Info. Overhead}
        \label{fig:f5_svhn_info_exp}
    \end{subfigure}
    \hfill
    \begin{subfigure}[b]{0.3\textwidth}
    \centering
    \includegraphics[height=4cm]{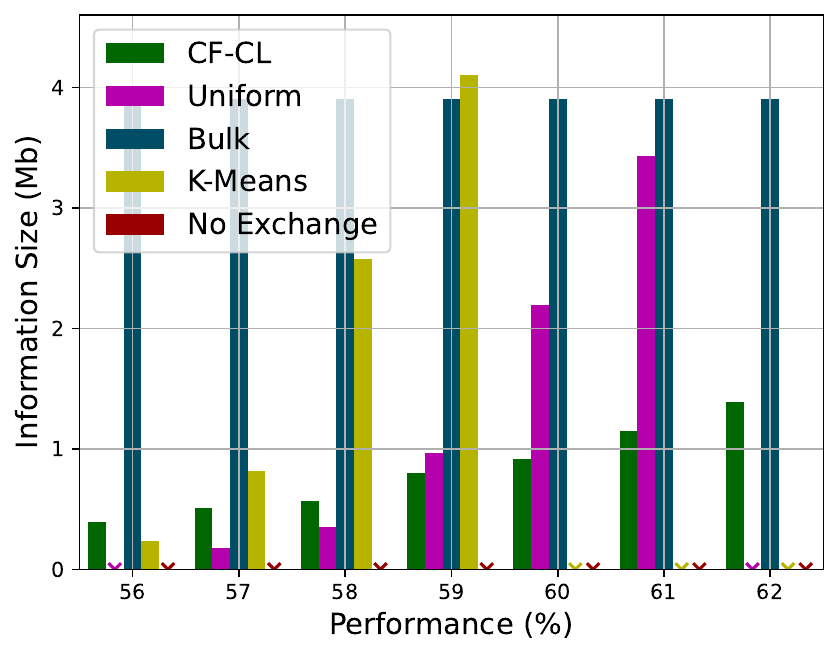}
        \caption{USPS Explicit Info. Overhead}
        \label{fig:f5_usps_info_exp}
    \end{subfigure}
    \hfill

    \begin{subfigure}[b]{0.3\textwidth}
    \centering
    \includegraphics[height=4cm]{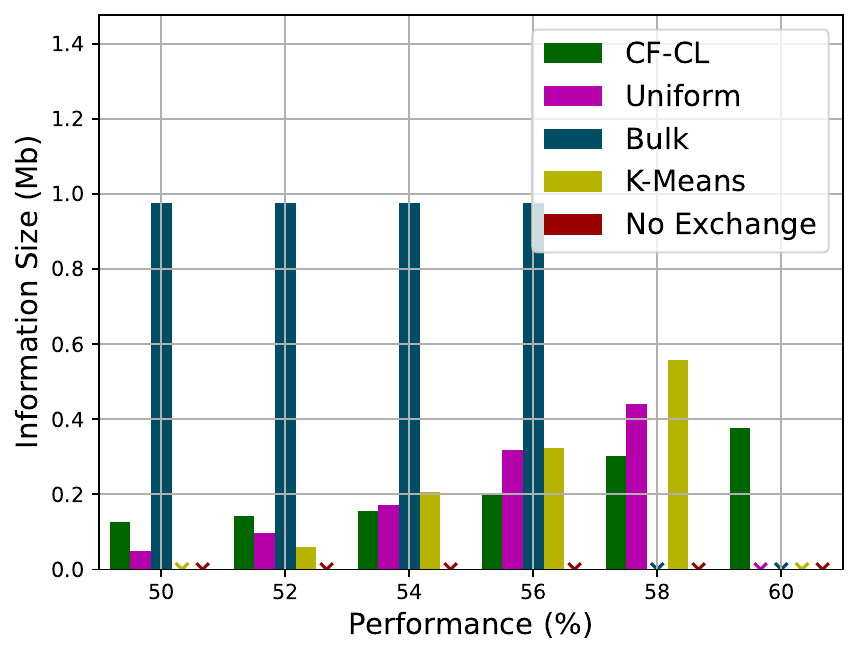}
        \caption{FMNIST Implicit Info. Overheads}
        \label{fig:f5_fmnist_info_imp}
    \end{subfigure}
    \hfill
    \begin{subfigure}[b]{0.3\textwidth}
    \centering
    \includegraphics[height=4cm]{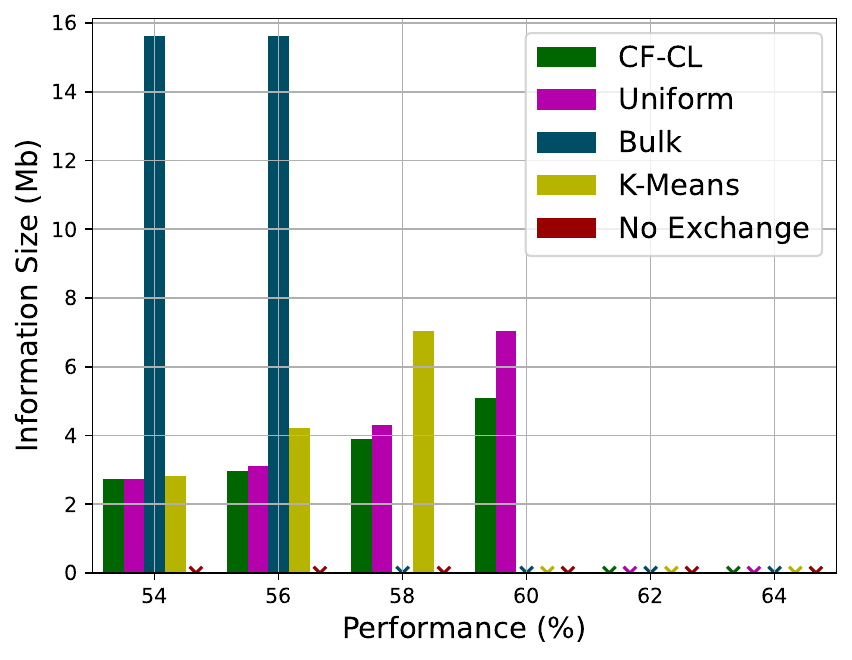}
        \caption{SVHN Implicit Info. Overhead}
        \label{fig:f5_svhn_info_imp}
    \end{subfigure}
    \hfill
    \begin{subfigure}[b]{0.3\textwidth}
    \centering
    \includegraphics[height=4cm]{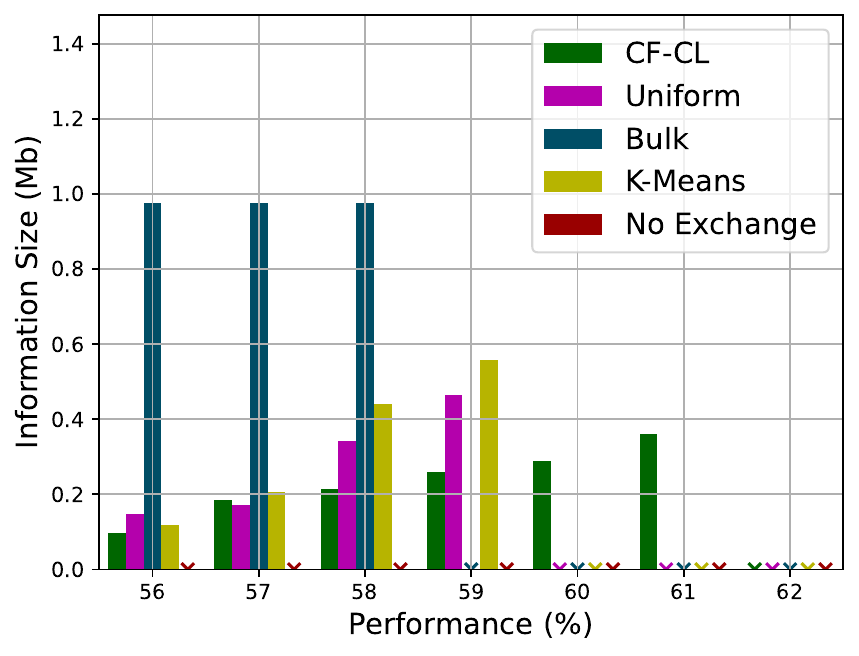}
        \caption{USPS Implicit Info. Overhead}
        \label{fig:f5_usps_info_imp}
    \end{subfigure}
    \hfill
    
    \begin{subfigure}[b]{0.3\textwidth}
    \centering
    \includegraphics[height=4cm]{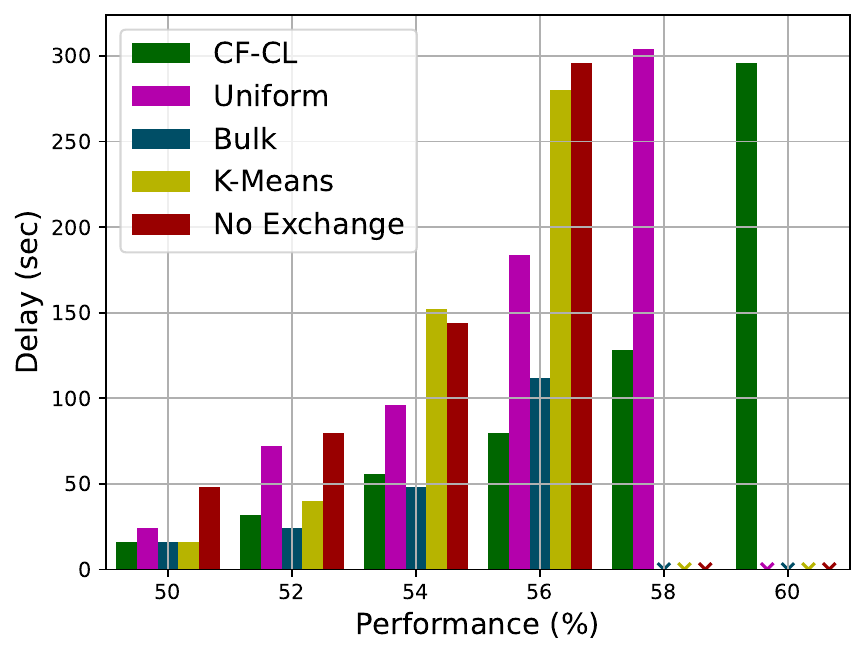}
        \caption{FMNIST Explicit Delays}
        \label{fig:f5_fmnist_delay_exp}
    \end{subfigure}
    \hfill
    \begin{subfigure}[b]{0.3\textwidth}
    \centering
    \includegraphics[height=4cm]{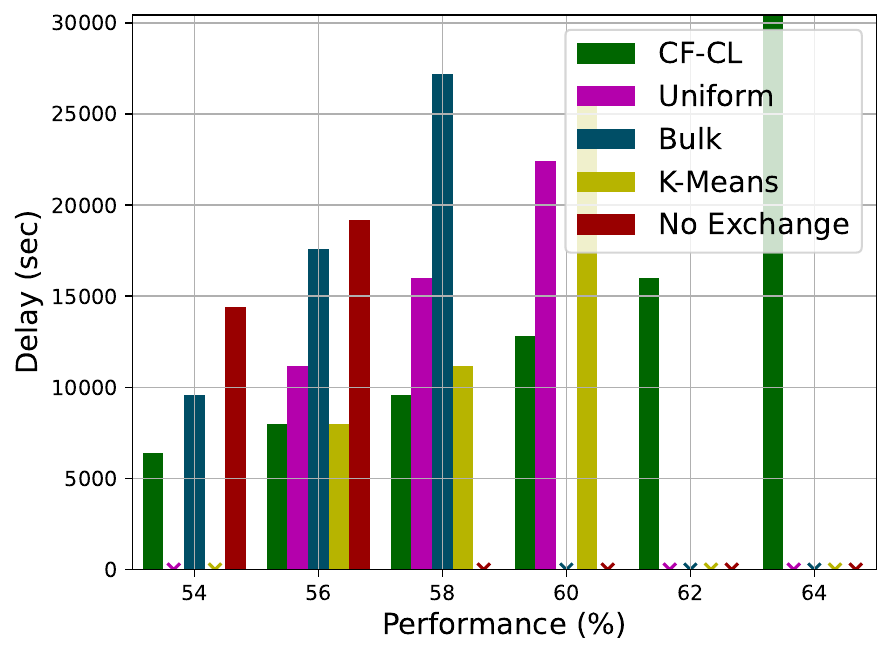}
        \caption{SVHN Explicit Delays}
        \label{fig:f5_svhn_delay_exp}
    \end{subfigure}
    \hfill
    \begin{subfigure}[b]{0.3\textwidth}
    \centering
    \includegraphics[height=4cm]{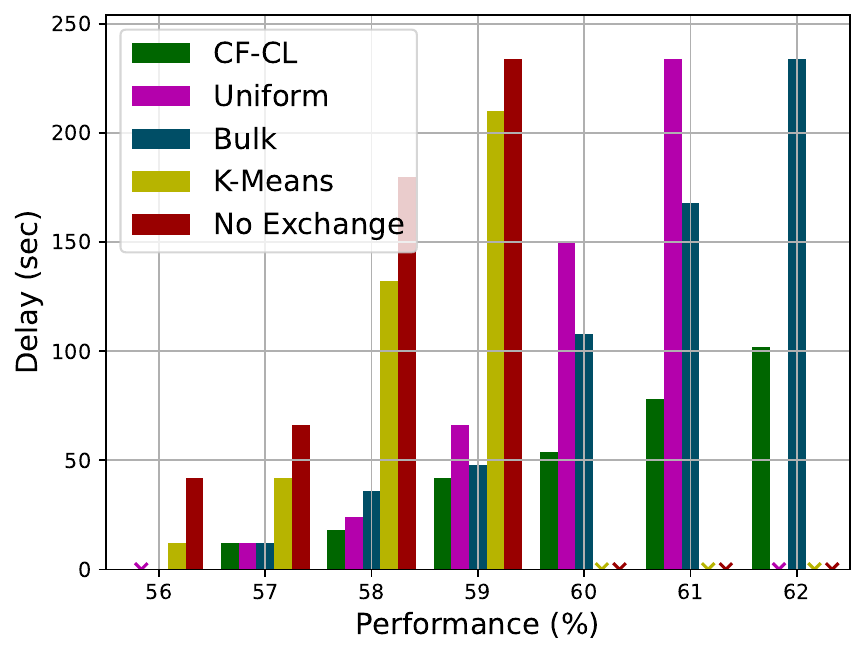}
        \caption{USPS Explicit Delays}
        \label{fig:f5_usps_delay_exp}
    \end{subfigure}
    \hfill

    \begin{subfigure}[b]{0.3\textwidth}
    \centering
    \includegraphics[height=4cm]{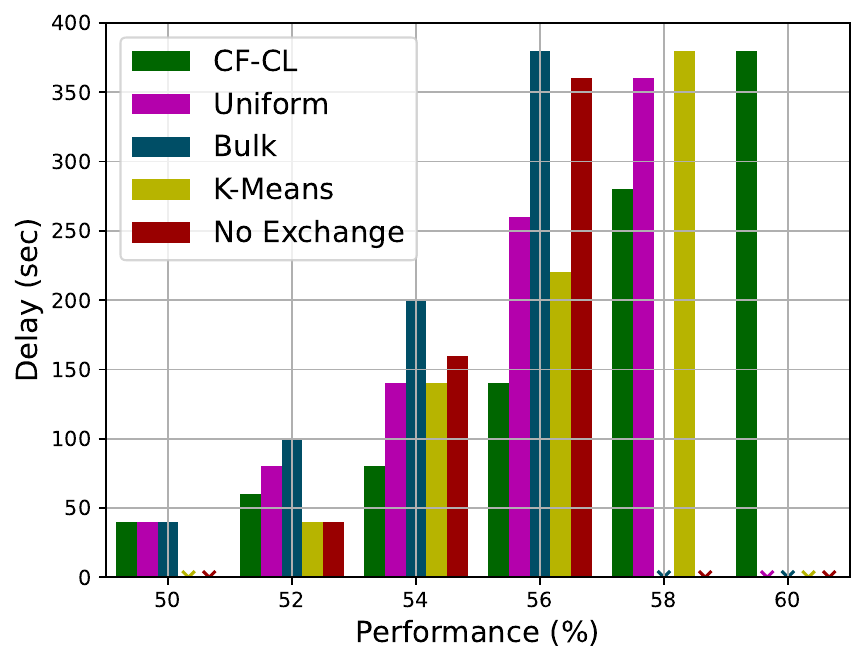}
        \caption{FMNIST Implicit Delays}
        \label{fig:f5_fmnist_delay_imp}
    \end{subfigure}
    \hfill
    \begin{subfigure}[b]{0.3\textwidth}
    \centering
    \includegraphics[height=4cm]{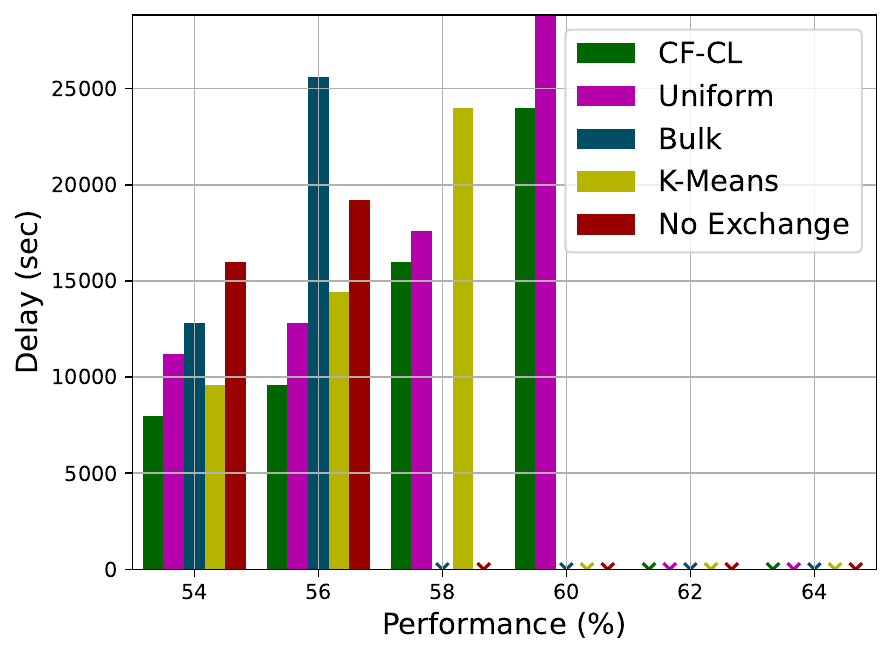}
        \caption{SVHN Implicit Delays}
        \label{fig:f5_svhn_delay_imp}
    \end{subfigure}
    \hfill
    \begin{subfigure}[b]{0.3\textwidth}
    \centering
    \includegraphics[height=4cm]{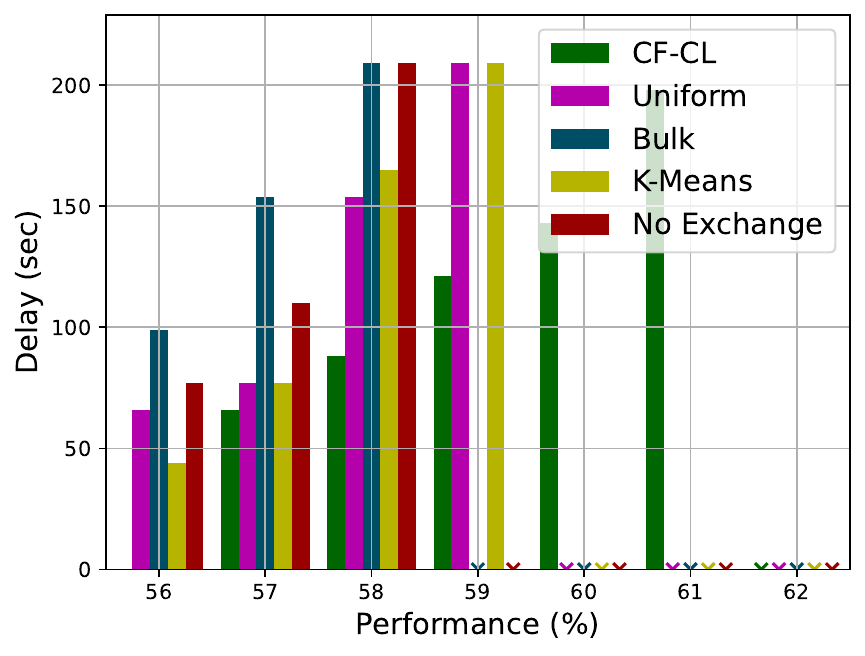}
        \caption{USPS Implicit Delays}
        \label{fig:f5_usps_delay_imp}
    \end{subfigure}
    \hfill

    \caption{{\tt{CF-CL}} incurs a smaller information overhead (\labelcref{fig:f5_fmnist_info_exp,fig:f5_svhn_info_exp,fig:f5_usps_info_exp,fig:f5_fmnist_info_imp,fig:f5_svhn_info_imp,fig:f5_usps_info_imp}) and is quicker (\labelcref{fig:f5_fmnist_delay_exp,fig:f5_svhn_delay_exp,fig:f5_usps_delay_exp,fig:f5_fmnist_delay_imp,fig:f5_svhn_delay_imp,fig:f5_usps_delay_imp}) to achieve performance benchmarks compared to baselines. Explicit exchange achieves benchmarks faster than implicit exchange at the cost of larger information overhead. ($\times$ indicates that the method was unable to reach the performance threshold.)}\label{fig:acc_delay_info}
%\vspace{-8mm}
\end{figure*}
%\fi
%\ali{In the legend of figures, you have ``Data" and ``Embs". I think you need to replace them with ``Explicit" and ``Implicit". Also, save the figures as pdf not png so they have a higher quality.}

\subsection{Comparison against Baselines}
    \textit{\textbf{Embedding Alignment:}} In Fig.~\ref{fig:emb_distances}, we consider the embeddings generated by conventional FL (Fig. \ref{fig:f7_left}), explicit {\tt CF-CL} (Fig. \ref{fig:f7_center}) and implicit {\tt CF-CL} (Fig. \ref{fig:f7_right}) 
    %\ali{This does not match the caption of the figure! Is the middle one with explicit or implicit?}\satya{changed}for the Fashion MNIST dataset \ali{, we consider XXX dataset} \satya{changed} 
    and calculate the Euclidian distance between all pairs of embeddings from all combinations of labels. We observe that smart data transfer in {\tt CF-CL} in both implicit and explicit cases leads to a latent space with more separated embeddings as compared to conventional FL, i.e., datapoints with distinct labels are further from each other as compared to those that have identical labels which appear on the diagonal of the heatmaps. While explicit {\tt CF-CL} can distinguish embeddings to a larger extent, the separation produced by both explicit and implicit cases results in a significant improvement in performance, as we will show in Fig. \ref{fig:acc_delay_info}.

    \textit{\textbf{Training Convergence Performance:}}
    In Fig. \labelcref{fig:f5_fmnist_accs_exp,fig:f5_svhn_accs_exp,fig:f5_usps_accs_exp,fig:f5_fmnist_accs_imp,fig:f5_svhn_accs_imp,fig:f5_usps_accs_imp}, we study the model training performance of explicit {\tt CF-CL}, implicit {\tt CF-CL} and baseline methods.
    Both explicit and implicit {\tt CF-CL} outperform respective baselines in terms of achieved model accuracy over training iterations, due to their ability to select information crucial to the alignment of local models. More specifically, the improvement over baselines validates a few aspects of {\tt CF-CL}'s design:
    (a) exchanging information that is both representative of the transmitter’s local dataset and important to the receiver (i.e., compared with K-means and Uniform),
    (b) selecting reserve information which accurately represents the local data/embedding distributions, and
    (c) adapting to the changes in local embedding distributions over the training duration.(i.e., compared to Bulk). We observe that in all cases, explicit {\tt CF-CL} converges faster than implicit {\tt CF-CL}. However, we will now show that this comes at the cost of significantly more information exchanged over the network. 
%\ali{We need to have a justification on why explicit is better in the bottom plot but implicit and explicit are on par in the top plot. Is it because of the complexity of the task where a more complex that would benefit from data (or embedding) exchange? this should be clear!}\satya{Changed. NOTE : Top row of Fig. 5 has been changed, I noticed that some of the results did not seem consistent, so I checked and reran the figure generation code.}
%\ali{What is a clear cut conclusion? If we just say implicit is the best, we would get asked then what is the benefit of explicit? You need to say sth like ``We will later show that this convergence speed accelerations comes with a higher latency (or sth else?) as compared to the explicit {\tt CF-CL} and baseline methods. }\satya{Changed}

    In Fig. \labelcref{fig:f5_fmnist_info_exp,fig:f5_svhn_info_exp,fig:f5_usps_info_exp,fig:f5_fmnist_info_imp,fig:f5_svhn_info_imp,fig:f5_usps_info_imp}, we measure the information overhead required to achieve performance milestones, i.e, to reach a particular testing accuracy percentage. 
%\ali{What plot are you talking about?}\satya{Changed}. 
    In the explicit case of {\tt{CF-CL}}, information exchanged are datapoints of size $784$ bytes for Fashion MNIST, $3072$ bytes for SVHN and $256$ bytes for USPS, considering each datapoint to be of the format specified above. For FMNIST and USPS in the implicit case of {\tt{CF-CL}}, the size of exchanged information is $64$ bytes per embeddings, as embeddings of size $16$ are used, with each element being a floating point value of size $4$ bytes. 
    %\cgb{Updated} 
    For SVHN, the embeddings are of size $256$, making the size of each embedding $1024$ bytes.We observe that both explicit {\tt{CF-CL}} and implicit {{\tt CF-CL}} utilize less information to meet target performance thresholds than the respective baselines, particularly as the threshold increases. This indicates that (a) despite the initial information overhead, the reserve data selection method of {\tt CF-CL} leads to more efficient model training by efficiently selecting representative information, and (b) periodically exchanging information according to the updated model training state is more efficient in improving performance. We also observe that the information overhead required by implicit {\tt{CF-CL}} is significantly smaller than explicit {\tt{CF-CL}}, owing to the smaller size of the exchanged information in implicit {\tt{CF-CL}}.

    \begin{figure*}[t!]
    \centering
		\includegraphics[width=0.9\linewidth,height=4.5cm]{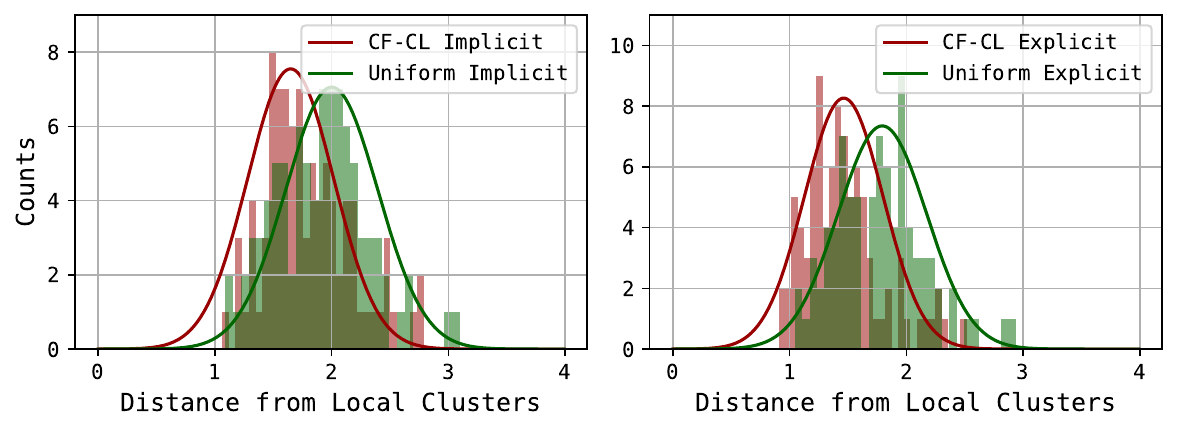}
		\caption{Proximity of received embeddings to local embeddings. In both Implicit (left) and Explicit (right) cases, {\tt{CF-CL}} selects information  that is harder to distinguish as negatives in the latent space compared to information selected by the uniform baseline, and are hence more important. 
%\ali{What is the left and right subplot? You need to specify and make that clear! See what I did in the caption of Fig. 9.}\satya{Done.}
}\label{fig:emb_dists}
%\vspace{-8mm}
\end{figure*}

    To further reveal the impact of faster convergence of {\tt CF-CL} on network resource savings, 
    we focus on the latency of model training as a performance metric in \labelcref{fig:f5_fmnist_delay_exp,fig:f5_svhn_delay_exp,fig:f5_usps_delay_exp,fig:f5_fmnist_delay_imp,fig:f5_svhn_delay_imp,fig:f5_usps_delay_imp}. We assume that transmission rate in D2D and uplink are $1$Mbits/sec with $32$ bits quantization applied on the model parameters ($45433$ for Fashion-MNIST, $26704$ for USPS and $1142352$ for SVHN) and $8$ on datapoints, which results in an uplink transmission delay of $45433 \times 32 / 10^6\approx 1.45$s  per model parameter exchange for Fashion MNIST, $26704 \times 32 / 10^6\approx 0.855$s for USPS and $1142352 \times 32 / 10^6\approx 36.566$s for SVHN. Fashion MNIST has a D2D delay of $28\times 28 \times 8 / 10^6 \approx 6.2$ms per data point exchange (each data point is a $28\times 28$ gray-scale image with each pixel taking $256$ values), USPS has a D2D delay of $16\times 16 \times 8 /  10^6 \approx 2$ms per data point exchange (each data point is a $16\times 16$ gray-scale image with each pixel taking $256$ values) and SVHN has a D2D delay of $3 \time 32\times 32 \times 8 /  10^6 \approx 24.5$ms per data point exchange, in case of the explicit exchange (each data point is a $32\times 32$ image with $3$ channels and each pixel taking $256$ values).
    Our models produce embeddings of size $16$ for both Fashion MNIST and USPS, thus resulting in a delay of $16 \times 32 / 10^6 \approx 0.512$ms per embedding in the implicit case (each embedding has a dimension $16$, each being a floating point number). Similarly, for implicit exchange using SVHN and embedding size $256$ results in a delay of $256 \times 32 / 10^6 \approx 8$ms. We also compute the extra computation time of {\tt CF-CL} (i.e., the K-means and importance calculations) and that of uniform sampling and incorporate that into delay computations. 
    %The right plot in Fig.~\ref{fig:3b} reveals significant delay savings that {\tt CF-CL} obtains\footnote{FedAvg is omitted from the plot due to its significantly lower performance.} upon reaching various accuracies ($\approx 18.7\%$ on average).
     In the last two rows of Fig. \ref{fig:acc_delay_info}, we observe that both variants of {\tt{CF-CL}} achieve performance milestones significantly faster than their respective baselines, indicating that (a) intelligent selection of reserve information and D2D information is crucial to delay reduction, (b) the delays incurred by D2D communication in {\tt{CF-CL}} are compensated for by the savings in device-to-server communication delays, and (c) {\tt{CF-CL}} cuts down on the upstream communication energy consumption required by reducing the number of global aggregations. 
%\ali{What plot are you talking about?}\satya{added}
    We also observe explicit {\tt{CF-CL}} being more efficient that implicit {\tt{CF-CL}} in terms of convergence speed for both datasets.   
%\ali{Compare explicit with implicit too! Say which one is better.}\satya{Done.}

    The above results indicate a tradeoff between latency and amount of information exchanged between explicit and implicit {\tt{CF-CL}}.
    %\blue{We also observe that {\tt{CF-CL}} requires significantly fewer global aggregations as compared to federated learning without D2D information exchange to achieve similar performance, indicating that the additional communication resources utilized by {\tt{CF-CL}} are more than compensated for by the reduction in device-to-server model communication cycles required.}
    We can conclude that implicit {\tt{CF-CL}} is best for bandwidth limited applications where minimal information exchange between devices is crucial, while explicit {\tt{CF-CL}} is best for latency critical applications.

   \textit{\textbf{Importance of Exchanged Information:}} 
%\ali{Figures should appear in the same order as you reference them! 8 is getting references before 6 and 7...}\satya{Changed. I still have some trouble arranging the figures such that they appear at the top of the page where the relevant paragraph is.}
   In Fig.~\ref{fig:emb_dists}, we calculate the average distance between the embeddings of received information $\mathcal{Z}_{j \rightarrow i}^t$ and the approximate local latent space occupied by $\{\phi_G^{\gamma T_a}(x)\}~\forall~ x \in \mathcal{D}$. The latent space is characterized by the centroids of $\{\phi_G^{\gamma T_a}(x)\}~\forall~ x \in \mathcal{D}$ after performing K-means. Intuitively, the closer the selected information is to local centroids in representation space, the more important it is. We construct a histogram of the average distance of each received embedding in $\mathcal{Z}_{j \rightarrow i}^t$ to observe the distribution of average distance over received information. We observe that {\tt{CF-CL}}, both implicit and explicit forms,  can select information which is more important to the receiving device, resulting in the peaks of the histograms for {\tt{CF-CL}} occurring at a distance closer to local data as compared to baseline methods. 
%\ali{This may not be correct! We are just talking about embeddings in Fig 8. Then, how we are addressing the ``explicit" CF-CL you  are mentioning above?}.\satya{In Fig 6, we are checking the distance between the rx information and local clusters in latent space. In explicit case, information is the datapoints, and they are passed through the local model to get embeddings, while in the implicit case, information is the embeddings themselves. I have clarified the caption to reflect that.}

\begin{figure*}[t!]
    \centering
    \begin{subfigure}[b]{0.49\textwidth}
            \centering
            \includegraphics[height=5cm]{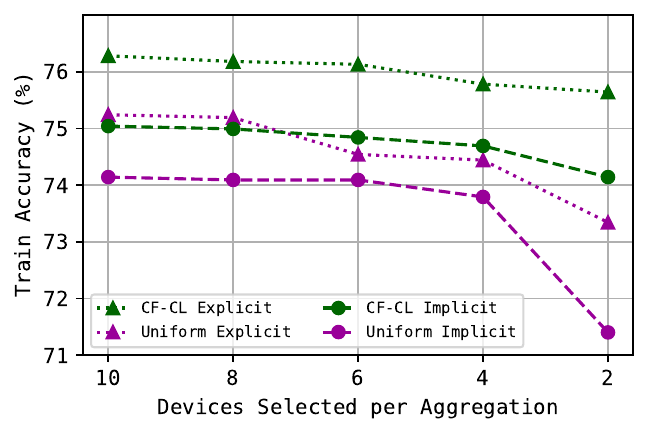}
                \caption{USPS}
                \label{fig:f7_usps}
    \end{subfigure}
    \hfill
    \begin{subfigure}[b]{0.49\textwidth}
            \centering
            \includegraphics[height=5cm]{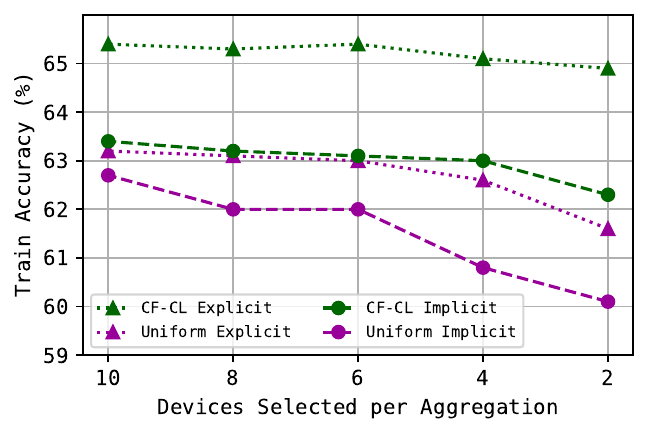}
                \caption{FMNIST}
                \label{fig:f7_fmnist}
    \end{subfigure}
    \vspace{-3mm}
    \caption{Change in number of devices selected per iteration on performance of {\tt{CF-CL}} against baselines for both datasets. {\tt{CF-CL}} is more resilient to a reduction in the number of local models being aggregated. 
%\ali{Either use ``client" or ``device" across the paper. Make sure everything is unified. I would suggest going with ``device".}\satya{Changed}
}
    \label{fig:agg_clients}
%\vspace{-8mm}
\end{figure*}

   \textit{\textbf{Number of Devices Selected for Aggregation:}} In Fig.~\ref{fig:agg_clients}, we observe the effect of changing the number of devices selected per aggregation on the performance of {\tt{CF-CL}} and compare it to baselines. In this experiment, local models are aggregated only from a subset of $n$ devices uniformly at random every $T_a$ time-steps, which is then broadcast to all devices. Both implicit and explicit data exchange methods for {\tt{CF-CL}} consistently outperform baselines. This indicates the ability of {\tt{CF-CL}} adapt to conditions where only a subset of devices can communicate with the server for aggregation. This indicates that {\tt{CF-CL}} is more resilient to a reduction in the number of local models being aggregated. This characteristic can be leveraged to conserve resources by reducing the frequency of communication of each device with the server. 

%\begin{figure*}
%		\centering
		%\includegraphics[width=0.48\linewidth,height=3cm]{images/knee_data.png}
		%\includegraphics[width=0.48\linewidth,height=3cm]{images/knee.png}
	%	\caption{Performance gain of {\tt{CF-CL}} against baselines relative to amount of information exchanged}\label{fig:knee}
%\end{figure*}	

\subsection{Impact of Variation in System Parameters}
   
%   This suggests optimal selection temperature as a promising future direction.

\begin{figure*}[t!]
        \centering
        \begin{subfigure}[b]{0.49\textwidth}
                \centering
                \includegraphics[height=5cm]{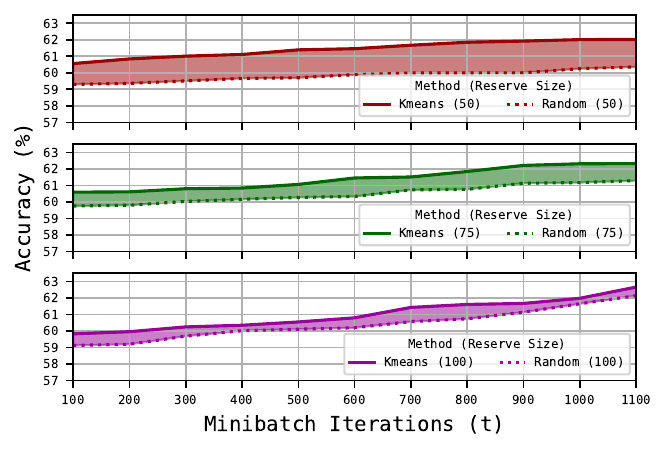}
                    \caption{Implicit {\tt CF-CL} Exchange}
                    \label{fig:f8_imp}
        \end{subfigure}
        \hfill
        \begin{subfigure}[b]{0.49\textwidth}
                \centering
                \includegraphics[height=5cm]{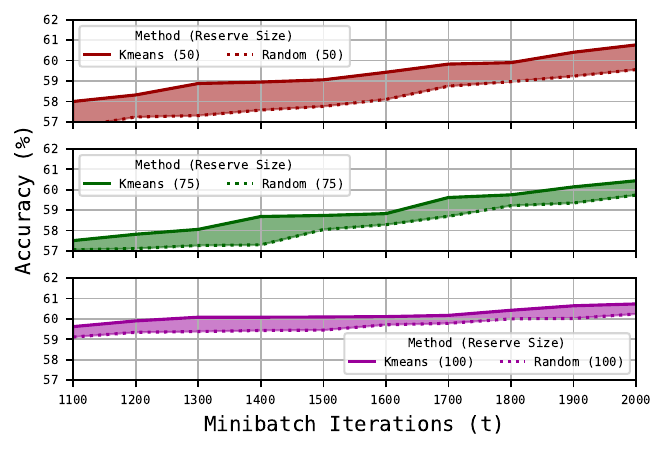}
                    \caption{Explicit {\tt CF-CL} Exchange}
                    \label{fig:f8_exp}
        \end{subfigure}
	\caption{Effects of using K-means and random sampling for reserve datapoints selection. Using K-means to select reserve datapoints results in better performance than selecting reserve data uniformly at random. }\label{fig:kmeans_comp}
    %\ali{What is the left and right subplot? You need to specify and make that clear! See what I did in the caption of Fig. 9.}\satya{Changed.}
%\vspace{-8mm}
\end{figure*}

     \textit{\textbf{Reserve Data Selection Methods:}} 
     %\ali{Change the x-axis of the three plots to ``Exchange Interval ($T_p$), Global Aggregation Index in which Data Exchange Starts, Global Aggregation Index in which Data Exchange Stops}\satya{Changed}
    In Fig.~\ref{fig:kmeans_comp}, we investigate the effect of using random sampling of $K^{\mathsf{Push}}_{i \rightarrow j}$, $\forall i,j$, data points as reserved datapoints vs. K-means based selection (Sec.~\ref{sec:PushKmeans}), in which device $i$ selects reserve information by running a K-means algorithm on local data with $K^{\mathsf{Push}}_{i \rightarrow j}$ clusters, under varying $K^{\mathsf{Push}}_{i \rightarrow j}$.
    From Fig.~\ref{fig:kmeans_comp}, performance of both methods of {\tt{CF-CL}} improve with selection of reserve data using K-Means. This is because K-means selects datapoints that best approximate the data distribution at each device. The effect of which is more significant in extreme cases, e.g., $K^{\mathsf{Push}}_{i \rightarrow j}=50$, $\forall i,j$, and diminishes as the allowable number of pushed data increases. This is because, as the number of reserve data increases, data selected at device $i$ by random sampling becomes more similar to the actual data distribution at device $i$.

\begin{figure*}[t!]
    \centering
        \begin{subfigure}[b]{0.49\textwidth}
                \centering
                \includegraphics[height=5cm]{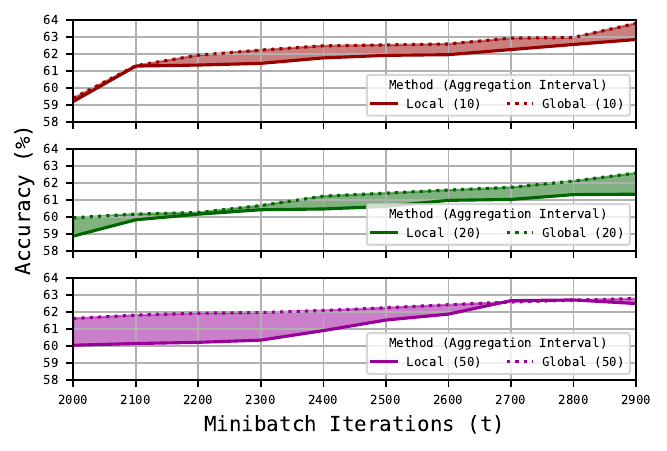}
                    \caption{Implicit {\tt CF-CL} Exchange}
                    \label{fig:f9_imp}
        \end{subfigure}
        \hfill
        \begin{subfigure}[b]{0.49\textwidth}
                \centering
                \includegraphics[height=5cm]{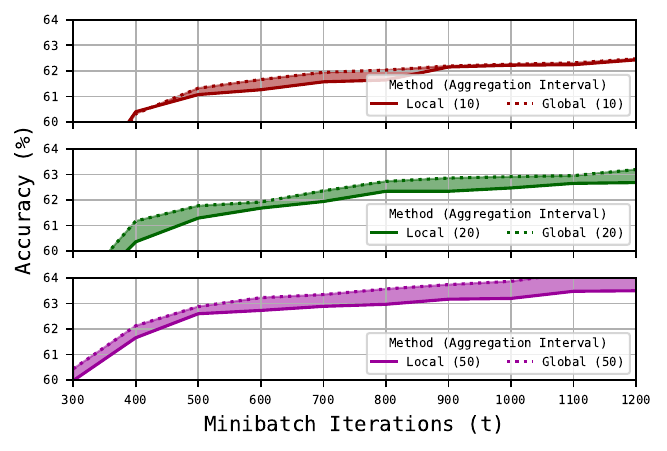}
                    \caption{Explicit {\tt CF-CL} Exchange}
                    \label{fig:f9_exp}
        \end{subfigure}
        %\vspace{-3mm}
        \hfill
		\caption{Variation in performance of {{\tt CF-CL}} when local and global models are used for importance calculations for both information exchange methods. {\tt{CF-CL}} is resilient to differences in local models at transmitter and receiver. }\label{fig:local_global}
%\vspace{-8mm}
\end{figure*}

    \textit{\textbf{Local vs. Global Models for Importance Calculation:}} 
    Fig.~\ref{fig:local_global} illustrates the impact of using either local or global models for importance calculations at the transmitter and receiver. For a constant exchange interval $T_p = 25$, when aggregation and exchange are asynchronous, the performance gap between using local model $\phi_i^{\tau T_p}$ and using latest global model $\phi_G^{\gamma T_a}$ where $\gamma T_a \leq \tau T_p < (\gamma+1) T_a$ for importance calculation increases as the aggregation interval $T_a$ increases, indicating that {\tt{CF-CL}} obtains performance gains even if models used for importance calculations are different. This shows that for {\tt{CF-CL}}, for the purposes of importance calculation, synchronization of local models at every exchange interval is not necessary for importance calculations, i.e, global aggregations are not needed at exchange intervals. This implies global knowledge is not necessary for importance calculation, which is desirable. We also observe that the explicit version of {\tt{CF-CL}} is more resilient to local model differences, this is consistent with the fact that the information exchanged is not a function of the model at the transmitter. 

\begin{figure*}[t!]
	\centering
        \begin{subfigure}[b]{0.49\textwidth}
                \centering
                \includegraphics[height=5cm]{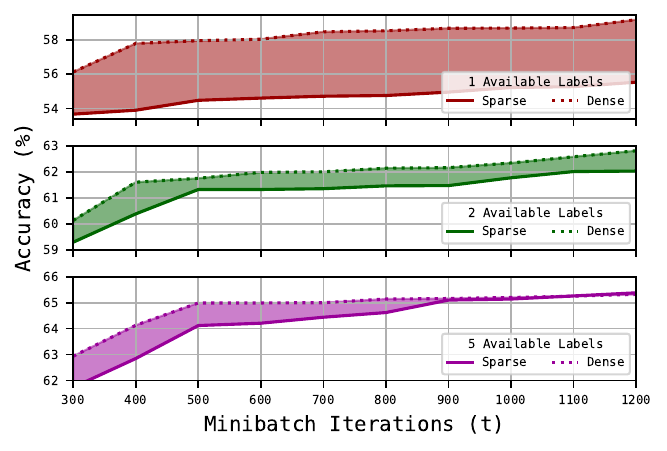}
                    \caption{FMNIST Explicit {\tt CF-CL} Exchange}
                    \label{fig:f10_fmnist_exp}
        \end{subfigure}
        \hfill
        \begin{subfigure}[b]{0.49\textwidth}
                \centering
                \includegraphics[height=5cm]{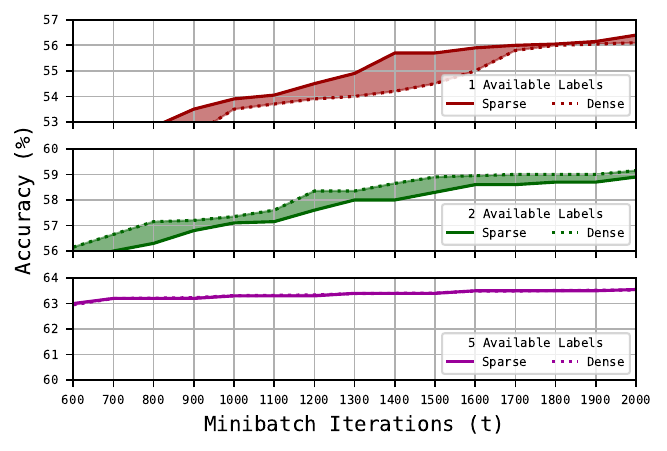}
                    \caption{FMNIST Implicit {\tt CF-CL} Exchange}
                    \label{fig:f10_fmnist_imp}
        \end{subfigure}
        \hfill
        \begin{subfigure}[b]{0.49\textwidth}
                \centering
                \includegraphics[height=5cm]{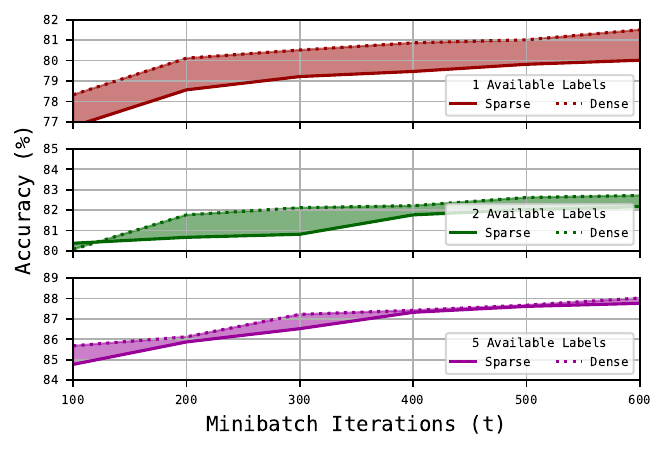}
                    \caption{USPS Explicit {\tt CF-CL} Exchange}
                    \label{fig:f10_usps_exp}
        \end{subfigure}
        \hfill
        \begin{subfigure}[b]{0.49\textwidth}
                \centering
                \includegraphics[height=5cm]{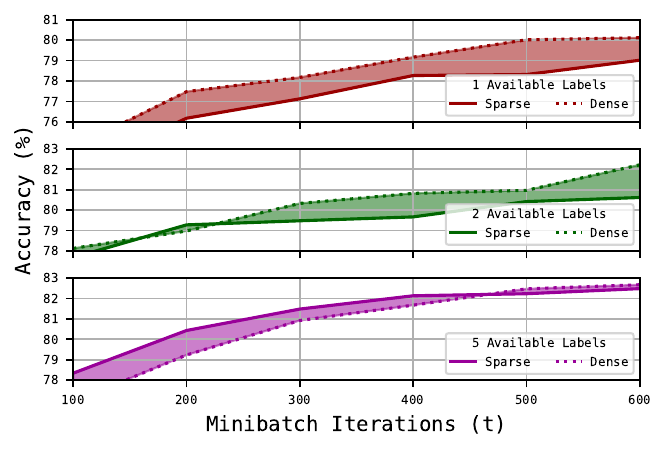}
                    \caption{USPS Implicit {\tt CF-CL} Exchange}
                    \label{fig:f10_usps_imp}
        \end{subfigure}
        
    %\includegraphics[width=0.48\linewidth,height=5cm]{images/graph_imp.pdf}
    %\includegraphics[width=0.48\linewidth,height=5cm]{images/graph_data.png}
    %\includegraphics[width=0.48\linewidth,height=5cm]{images/graph_usps_data.pdf}
    %\includegraphics[width=0.48\linewidth,height=5cm]{images/graph_usps.pdf}
	%\vspace{-1.5mm}
	\caption{Effect of connectivity of devices under varying conditions of non-i.i.d. data. Higher connectivity improves the performance of {{\tt CF-CL}} in conditions where locally available data is not diverse.}\label{fig:conn}
%	\vspace{-7mm}
%\vspace{-8mm}
\end{figure*}

    \textit{\textbf{Local Data Availability and Device Connectivity:}} Fig.~\ref{fig:conn} shows different scenarios of non-i.i.d data with varying connectivity between devices. We consider cases where the average node degree in the communication graph $\mathcal{G}$ is $2$ for a sparsely connected graph and $8$ for a densely connected graph.  We vary the number of labels in each device's local dataset and show that as the local data distributions become more non-i.i.d. (i.e., fewer labels per device), the speed of convergence of the methods drop due to more biased local models. In such cases, higher connectivity significantly improves the performance, as a higher connectivity allows for the exposure of local datasets to a more diverse set of information resulting in less biased local models. We also find that in Fig. \ref{fig:f10_fmnist_exp} and Fig. \ref{fig:f10_usps_exp}, the performance on the FMNIST dataset improves significantly more than on the USPS dataset, as classification using FMNIST is a more complex learning task, and thus affected by non-i.i.d to a larger extent. We also observe that in the case of implicit information exchange, once the number of labels per device become large enough, the performance of our method for sparse and densely connected graphs is either equivalent as in Fig. \ref{fig:f10_fmnist_imp} or our method performs better in a sparsely connected graph as in Fig. \ref{fig:f10_usps_imp}. This is because the implicit form of {\tt{CF-CL}} employs regularization to incorporate implicit information into training, as described in \eqref{eq:triplet_loss_reg}. A densely connected graph of devices, where each device has local access to a large number of labels may result in the exchange of redundant information, resulting in a slower convergence rate, especially in the earlier stages of training. 

\vspace{-1.5mm}
\section{Conclusion and Future Work}
\noindent In this paper, we proposed Cooperative Federated unsupervised Contrastive Learning ({\tt CF-CL}). In {\tt CF-CL}, devices learn local representations of unlabeled data and engage in cooperative smart explicit or implicit information push-pull to eliminate the local model bias. In case of explicit information exchange, we proposed an efficient randomized data importance estimation and subsequently developed a two-staged probabilistic data sampling scheme across the devices. In case of implicit information exchange, we  designed a probabilistic embedding exchange scheme and modified the definition of triplet loss function to use the exchanged embeddings while taking into account for their freshness.
Through numerical simulations, we studied the model training behavior of {\tt CF-CL} and showed that it outperforms the baseline methods in terms of accuracy and efficiency.
Future work can extend {\tt{CF-CL}} from single-hop information sharing to multi-hop information sharing and D2D link formation under network constraints. Such a system will require a more nuanced method of information importance estimation to take into account first hop neighbors, as well as higher order neighbours in the D2D graph.
%Assigning non-uniform costs to information transfer based on network conditions will further increase {\tt{CF-CL}}'s practicality.

% \pagebreak

\vspace{-2.5mm}
\bibliographystyle{IEEEtran}
\bibliography{fcl-refs.bib}

\end{document}